\documentclass[11pt]{article}

\usepackage[preprint]{acl}

\usepackage{times}
\usepackage{latexsym}

\usepackage{hyperref}
\usepackage{url}
\usepackage{booktabs} 
\usepackage{colortbl}
\usepackage{amsmath}
\usepackage{amssymb}
\usepackage{mathtools}
\usepackage{amsthm}
\usepackage{titling}
\usepackage{adjustbox} 
\usepackage{multirow}  
\usepackage{makecell}  
\usepackage{booktabs}  
\usepackage{caption}
\usepackage{adjustbox}
\usepackage{fontawesome}

\usepackage{lipsum}

\usepackage{titletoc} 
\usepackage{hyperref} 

\usepackage[most]{tcolorbox} 
\definecolor{skyblue}{RGB}{203, 221, 245}
\definecolor{forestgreen}{HTML}{228B22}
\definecolor{coralred}{HTML}{FF6F61}
\newcommand{\skyblue}{\rowcolor{skyblue}}
\usepackage{wrapfig} 

\usepackage{pifont}

\newcommand{\cmark}{{\color{forestgreen}{\faCheckCircle}}}
\newcommand{\xmark}{{\color{coralred}{\faTimesCircle}}}

\usepackage[T1]{fontenc}
\usepackage[utf8]{inputenc} 
\DeclareUnicodeCharacter{2261}{$\equiv$}

\usepackage[utf8]{inputenc}

\usepackage{microtype}

\usepackage{inconsolata}

\usepackage{graphicx}

\title{\textit{\includegraphics[width=26pt,height=18pt]{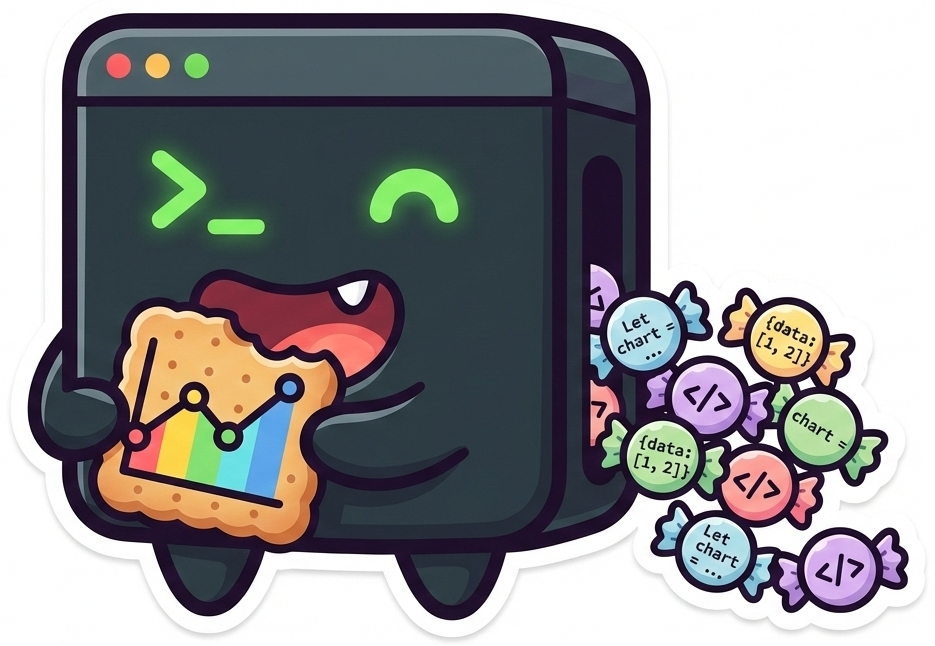}~\texttt{RealChart2Code}}: Advancing Chart-to-Code \\ Generation with Real Data and Multi-Task Evaluation}

\author{
 \textbf{Jiajun Zhang\textsuperscript{1,5}\thanks{Equal contribution.}}\quad\!\!\!\!
 \textbf{Yuying Li\textsuperscript{2}$^*$}\quad\!\!\!\!
 \textbf{Zhixun Li\textsuperscript{3}$^*$}\quad\!\!\!\!
 \textbf{Xingyu Guo\textsuperscript{4,5}}\quad\!\!\!\!
 \textbf{Jingzhuo Wu\textsuperscript{6}}\quad\!\!\!\!
 \\
 \textbf{Leqi Zheng\textsuperscript{2}}\quad\!\!\!\!
 \textbf{Yiran Yang\textsuperscript{7}}\quad\!\!\!\!
 \textbf{Jianke Zhang\textsuperscript{2}}\quad\!\!\!\!
 \textbf{Qingbin Li\textsuperscript{4}}\quad\!\!\!\!
 \textbf{Shannan Yan\textsuperscript{2}}\quad\!\!\!\!
 \textbf{Zhetong Li\textsuperscript{8}}\quad\!\!\!\!
 \\
 \textbf{Changguo Jia\textsuperscript{9}}\quad\!\!\!\!
 \textbf{Junfei Wu\textsuperscript{4,5}}\quad\!\!\!\!
 \textbf{Zilei Wang\textsuperscript{1}}\quad\!\!\!\!
 \textbf{Qiang Liu\textsuperscript{4,5}}\quad\!\!\!\!
 \textbf{Liang Wang\textsuperscript{4,5}}
 \\
 \textsuperscript{1}USTC\quad\!\!
 \textsuperscript{2}THU\quad\!\!
 \textsuperscript{3}CUHK\quad\!\!
 \textsuperscript{4}UCAS\quad\!\!
 \textsuperscript{5}CASIA\quad\!\!
 \textsuperscript{6}BNU\quad\!\!
 \textsuperscript{7}BUPT\quad\!\!
 \textsuperscript{8}BIT\quad\!\!
 \textsuperscript{9}PKU
 \\
 {\faEnvelope} \texttt{zhangjiajun519@gmail.com} \\[0.2em]
 \faDatabase\ \textbf{Dataset:} \url{https://huggingface.co/datasets/zjj1233/RealChart2Code}
}

\begin{document}
\maketitle
\begin{abstract}

Vision-Language Models (VLMs) have demonstrated impressive capabilities in code generation across various domains. However, their ability to replicate complex, multi-panel visualizations from real-world data remains largely unassessed. To address this gap, we introduce \textbf{\texttt{RealChart2Code}}, a new large-scale benchmark with over 2,800 instances grounded in authentic datasets and featuring tasks with clear analytical intent. Crucially, it is the first benchmark to systematically evaluate chart generation from large-scale raw data and assess iterative code refinement in a multi-turn conversational setting.
Our comprehensive evaluation of 14 leading VLMs on \texttt{RealChart2Code} reveals significant performance degradation compared to simpler benchmarks, highlighting their struggles with complex plot structures and authentic data. Our analysis uncovers a substantial performance gap between proprietary and open-weight models and confirms that even state-of-the-art VLMs often fail to accurately replicate intricate, multi-panel charts. These findings provide valuable insights into the current limitations of VLMs and guide future research directions. 
We release the benchmark and code at \url{https://github.com/Speakn0w/RealChart2Code}.

\end{abstract}

\section{Introduction}
\begin{figure*}[tp]
\centering
\includegraphics[width=1.0\linewidth]{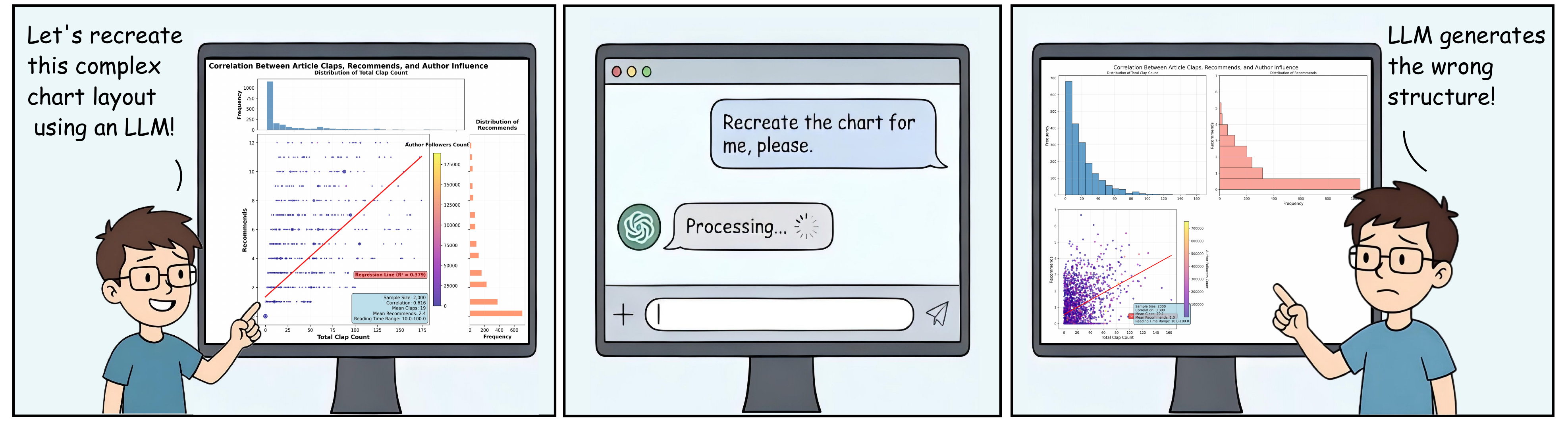}
\vspace{-1em}
\caption{A real-world example illustrating the limitations of LLMs on complex chart-to-code tasks. When presented with a sophisticated request, the model generates a low-quality output and struggles to make effective improvements during the subsequent refinement process.}
\label{fig:intro_fig}

\vspace{-0.5em}
\end{figure*}

Recent advancements in AI research have demonstrated the powerful code generation capabilities of LLMs \citep{gpt4,gpt5,claude,gemini1.5,code_llama,qwen25coder,codestral,kimi_k2,cao2026qwen3,team2025qwen3}, which have solved coding challenges in domains such as software engineering~\citep{swe-bench,swe-flow,swe-gym, shum2025swe}, code completion~\cite{crosscodeeval, execrepobench, safim, zhang2026completion}, and algorithmic problem-solving ~\citep{humaneval,bigcodebench,livecodebench}. Chart-to-code generation is another prominent application area, where the goal is to reproduce the visualization code from an image. This capability fulfills a frequent and practical user need by enabling users to recover the underlying visualization logic from static images, which is especially valuable when the original code is unavailable and the chart needs to be edited, extended, or reused in different contexts. However, while current VLMs excel at creating simple, single-panel charts, they struggle to generate plots with multiple subplots and intricate composite layouts, especially when derived from large, complex structured data. As illustrated in Figure \ref{fig:intro_fig}, a state-of-the-art model fails to accurately replicate the intended multi-plot structure.

\begin{table}[t]
\centering

\setlength{\tabcolsep}{3pt}
\resizebox{1\linewidth}{!}{
\begin{tabular}{l ccccc}
\toprule
\textbf{Benchmark}  & \textbf{Real Data} & \textbf{Complex} & \textbf{Interactive} &\textbf{Multi Tasks} \\
\midrule
Plot2Code      & \xmark & \xmark & \xmark & \xmark \\
Design2Code & \xmark & \xmark & \xmark & \xmark \\
ChartMimic     & \xmark & \xmark & \xmark &\cmark \\
\rowcolor{gray!15}
\textbf{Ours}  & \cmark & \cmark & \cmark &\cmark \\
\bottomrule
\end{tabular}}
\vspace{-0.5em}
\caption{Comparison of \texttt{RealChart2Code} with existing chart-to-code benchmarks.}
\label{tab:benchmark-comparison-singlecol-alt}
\vspace{-1em}
\end{table}

Prior benchmarks for chart-to-code generation have primarily focused on simple chart types and single-panel layouts. They often rely on either pre-existing chart-code pairs from the internet, which pose a risk of data leakage, or on synthetic data created to replicate figures from scientific papers (e.g., Plot2Code~\citep{plot2code}, ChartMimic~\citep{chartmimic}). Furthermore, they lack metrics for evaluating a model's ability to refine code in multi-turn conversation. With the rapid advancement of LLMs, such benchmarks are no longer sufficient for evaluating a model's ability to handle chart-to-code tasks involving complex, real-world data and intricate plot structures.

To systematically evaluate these capabilities, we introduce \texttt{RealChart2Code}, a new large-scale benchmark comprising 2896 instances. \texttt{RealChart2Code} is distinguished from prior work in four key aspects, as illustrated in Table \ref{tab:benchmark-comparison-singlecol-alt}. \ding{182} First, it is grounded in realistic visualization scenarios and utilizes authentic datasets, in contrast to benchmarks that rely on synthetic data or arbitrary constructions. \ding{183} Second, it introduces a significantly higher level of complexity by incorporating intricate chart structures and a diverse range of chart types. \ding{184} Third, it features an interactive chart-to-code framework that simulates real-world development workflows. \ding{185} Finally, it incorporates three challenging tasks designed to comprehensively evaluate model capabilities in generating complex visualizations, understanding chart semantics, and modifying plots.
Specifically, we construct the benchmark by rigorously filtering high-quality datasets from Kaggle~\citep{kaggle}, manually designing complex visualization tasks, and implementing the corresponding ground-truth code. Furthermore, we construct realistic chart refinement contexts by manually designing errors and correction instructions, ultimately yielding a comprehensive benchmark grounded in authentic development scenarios.


We evaluate 14 prominent VLMs on the \texttt{RealChart2Code} benchmark, including 5 proprietary and 9 open-weight models. We observe that most models that perform well on simple benchmarks fail to achieve comparable performance on \texttt{RealChart2Code}, primarily due to difficulties in handling complex chart structures and large-scale, authentic data. To validate our quantitative results, we conduct a human evaluation that manually inspects the correctness and fidelity of the generated visualizations. A subsequent correlation analysis (\S~\ref{correlation_analysis}) demonstrates a strong correlation between our multi-level metrics and human judgments.
Finally, we perform extensive quantitative analysis and qualitative case studies (\S\ref{sec:discussion}) on model performance across multiple benchmarks (\S\ref{sec:cross_benchmark_analysis}). This analysis reveals key similarities and differences in model capabilities across tasks of varying difficulty and types, providing valuable insights to guide future research.
\section{Related Works}
\subsection{Code Generation}
Recent advances in Large Language Models (LLMs), including general-purpose models (e.g., GPT \citep{gpt4}, Claude \citep{claude}, Gemini \citep{gemini1.5}) and specialized code models (e.g., Qwen-Coder \citep{qwen25coder}, DeepSeek-Coder \citep{deepseek_coder}, Codestral \citep{codestral}), have demonstrated powerful coding capabilities \citep{deepseekv3,code_llama,kimi_k2,glm_4.5}. While conventional tasks like algorithmic problem-solving \citep{humaneval, bigcodebench} and software engineering \citep{swe-bench,swe-flow} are evaluated on functional correctness \citep{pass@1}, this paper focuses on data visualization, a domain where generated code must produce a visually accurate output, a requirement shared by front-end design \citep{web-bench,webgen-bench,chen2025interactscience} and SVG generation \citep{llm4svg}.

\begin{figure*}[!t]
\centering
\includegraphics[width=\linewidth]{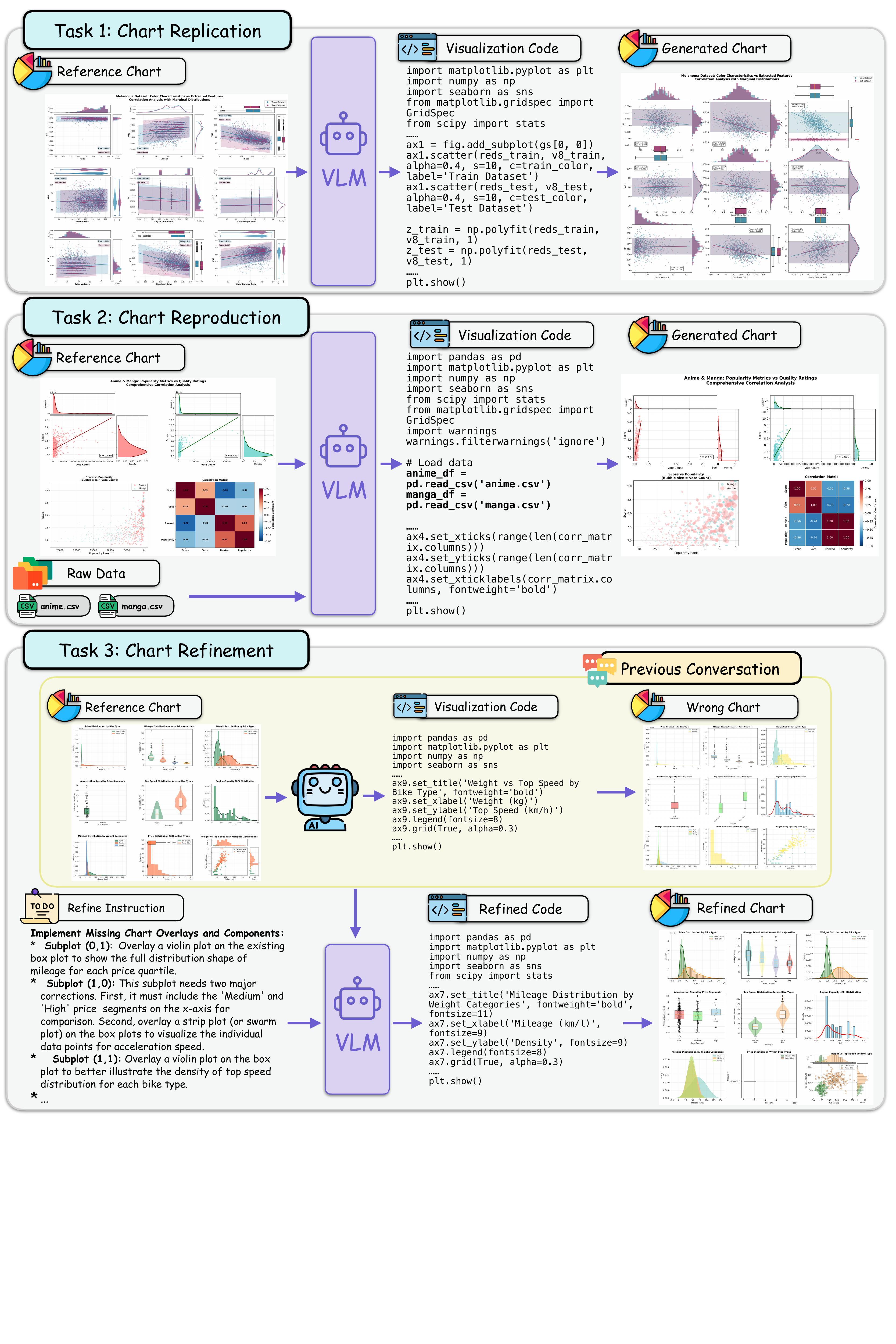}
\vspace{-1em}
\caption{The three core tasks of the RealChart2Code benchmark. \textbf{Chart Replication} is the fundamental chart-to-code task. \textbf{Chart Reproduction} adds the challenge of using provided raw data files. \textbf{Chart Refinement} introduces a conversational component, where the model must debug and modify code to fix errors according to user feedback.}
\label{fig:task}
\vspace{-0.5em}
\end{figure*}

\subsection{Data Visualization}
Prior LLM-based data visualization research spans three main areas. The first, chart understanding, focuses on interpreting visual information from plots for tasks like question answering or summary generation \citep{li2024mmsci,zeng2024advancing, ChartSumm, Chart2Text, chartreasoner, zhang2025tablellm, ma2024spreadsheetbench, zheng2026should, li2025capgeo}. The second, Text-to-Visualization (Text2Vis), concerns generating visualization specifications or code from natural language descriptions \citep{nvbench2.0, pandasbench, viscoder, plotcraft}. The third, Chart-to-Code (Chart2Code), involves reverse-engineering a visualization by generating the code required to replicate it \citep{plot2code, chartmimic, chartcoder}. However, existing benchmarks in this domain predominantly feature simple, single-panel plots, which are insufficient for evaluating an LLM's ability to handle complex layouts and high information density. To address this critical gap in chart-to-code evaluation, we introduce RealChart2Code, a benchmark specifically designed to assess performance on intricate, multi-panel charts derived from real-world data.

\section{\texttt{RealChart2Code}}

\begin{figure*}[tp]
\centering
\includegraphics[width=0.95\linewidth]{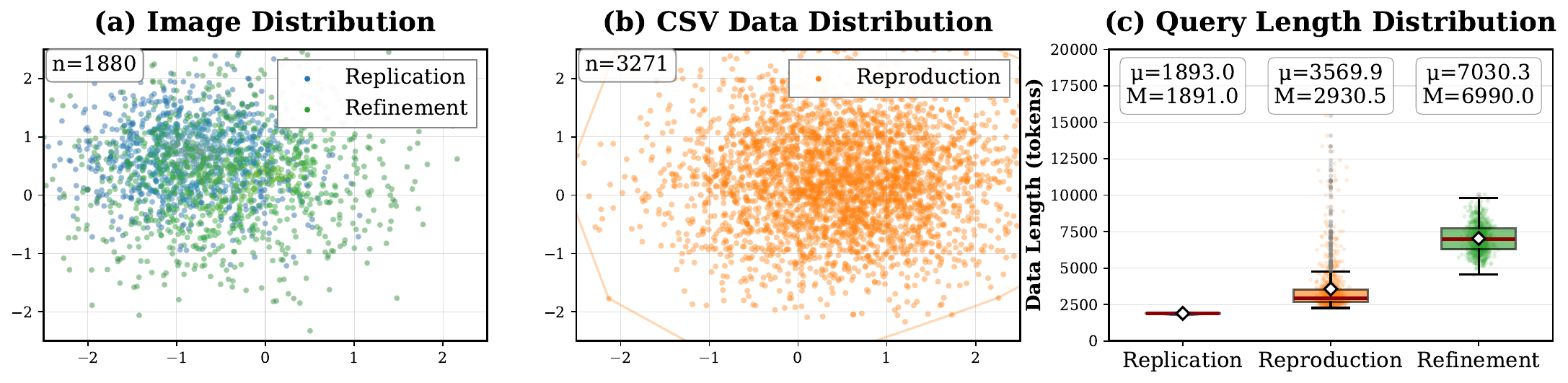}
\caption{Dataset distribution across tasks. (a) Image distribution. (b) CSV data distribution. (c) Data length distribution with median (M) and mean ($\mu$) in tokens.}
\label{fig:distribution}
\end{figure*}

\subsection{Task Definition}
We define the chart-to-code task as a conditional code generation problem. Formally, given a source chart image $V$ and an accompanying prompt $P$, a LLM, denoted by $\mathcal{F(\cdot)}$, must generate an executable code snippet $C$. This code must render a visualization that accurately reproduces the visual and structural elements of $V$ while adhering to any requirements in $P$. The task is formulated as $C = \mathcal{F}(V,P)$. The \texttt{RealChart2Code} benchmark evaluates models on three distinct variants of this core task, illustrated in Figure~\ref{fig:task}: \textbf{(1) Chart Replication}, is the fundamental chart-to-code task where the model must reverse-engineer the visualization from the image alone, measuring its core visual-to-code translation ability. \textbf{(2) Chart Reproduction}, provides the model with the chart image, raw data, and metadata, assessing its capability to generate the correct plot using large-scale, real-world data sources. \textbf{(3) Chart Refinement}, which requires the model to correct a chart with predefined errors through a multi-turn dialogue, assessing its ability to perform iterative debugging based on user instructions.

\subsection{Benchmark Coverage Analysis}
\paragraph{Chart Types}
The chart data in \texttt{RealChart2Code} can be classified from two perspectives: visualization intent and chart type. Our taxonomy includes seven high-level intent categories and 50 distinct plot types. Crucially, all visualizations in the benchmark are designed to be complex, featuring composite charts or intricate multi-panel layouts. As a result, a single plot type label is often insufficient to describe a given instance. Detailed examples of these categories and complex layouts are provided in Appendix \ref{app-sub:chart_type}.

\paragraph{Dataset Distribution}
\texttt{RealChart2Code} covers diverse thematic topics across eight high-level domains: Finance, Industry, Health, Research, Society, Media, Technology, and Environment. These domains are further divided into 35 fine-grained sub-topics, ensuring broad applicability to real-world scenarios.
Figure~\ref{fig:distribution} illustrates the distribution of chart images and CSV data across all three tasks using CLIP~\citep{clip} and t-SNE~\citep{t-SNE}. As shown in Figure~\ref{fig:distribution}(a) and (b), both distributions are widely dispersed across the feature space, indicating substantial diversity in visual styles, layouts, and data characteristics across Chart Replication, Chart Reproduction, and Chart Refinement tasks. Figure~\ref{fig:distribution}(c) shows the data length distribution, reflecting the complexity of real-world datasets. This comprehensive coverage ensures that \texttt{RealChart2Code} challenges models with diverse chart types and data patterns.

\subsection{Data Curation Process}
The construction of \texttt{RealChart2Code} follows a four-stage pipeline: (1) Data Collection and Filtering, (2) Visualization Task Design, (3) Code Implementation, and (4) Error Injection. We describe each stage in detail below. Further details and strict quality control measures are provided in Appendix \ref{app:data_curation}.
\paragraph{Data Collection and Filtering}
We collect open-source datasets from Kaggle. Our use of these datasets is strictly for scientific research; other uses fall under their original licenses. Our process involved a two-stage filtering pipeline. First, we performed an initial screening of over 8,000 datasets, which collectively contained more than 100,000 files and 30 billion data rows. This screening was based on community metrics such as vote counts, download counts, and usability ratings. From this initial pool, we conducted a second, more rigorous filtering stage to select 1,036 high-quality datasets suitable for our benchmark's task and chart construction. The final curated collection for \texttt{RealChart2Code} contains 3,271 raw data files, with approximately 860 million rows in total.

\paragraph{Visualization Task Design}
Using the curated datasets, we designed 1,016 unique and complex visualizations. Each of these visualizations serves as the basis for two distinct tasks: (1) Chart Replication, where the model receives only the chart image, and (2) Chart Reproduction, where the model is also provided with the corresponding raw data. This dual-task structure results in 2,032 instances across these two categories. Every visualization was designed to be contextually relevant to its source dataset, ensuring practical, real-world meaning. To guarantee task diversity and complexity, the design process was guided by a taxonomy of 7 high-level visualization intents and 50 distinct chart types.

\paragraph{Ground-Truth Code Implementation}
For each task, our in-house team of five expert Python developers implemented the ground-truth code using Matplotlib and its associated libraries. This code serves as the reference solution that models must replicate. The sandboxed execution environment used for evaluation is detailed in Appendix~\ref{app-sub:envirnment}.

\paragraph{Error Injection}
To create the Chart Refinement tasks, we manually injected errors into a subset of the ground-truth charts. These intentionally flawed charts serve as the starting point for a multi-turn dialogue. The types of errors are diverse, including incorrect chart types, data mapping errors, element overlap and other common errors. Through this process, we constructed 864 Chart Refinement tasks. In total, the \texttt{RealChart2Code} benchmark consists of 2,896 instances, comprising 1,016 for Chart Replication, 1,016 for Chart Reproduction, and the 864 for Chart Refinement.

\begin{table*}[!t]

\small
\centering
\begin{adjustbox}{width=\linewidth}
    \begin{tabular}{l|cc|cc|cc|c}
    \toprule
    \multirow{2}*{\makecell*[l]{Model}} 
    & \multicolumn{2}{c|}{\makecell*[c]{Chart Replication}}
    & \multicolumn{2}{c|}{\makecell*[c]{Chart Reproduction}} 
    & \multicolumn{2}{c|}{\makecell*[c]{Chart Refinement}}
    & \multirow{2}*{\makecell*[c]{AVG score}} \\
    & Pass (\%) & Score & Pass (\%) & Score & Pass (\%) & Score & \\
    \midrule

    \skyblue\multicolumn{8}{c}{\textit{Closed-source LLMs}}\\
    \cmidrule{1-8}
    Claude-4.5-Sonnet ~\citep{claude} & \underline{85.8} & 7.1 & \underline{73.7} & 5.7 & \underline{89.0} & 8.0 & 7.0 \\
    Claude-4.5-Opus~\citep{claude} & \textbf{87.7} & \underline{7.8} & \textbf{86.1} & \textbf{7.4} & \textbf{91.2} & \textbf{9.4} & \textbf{8.2} \\
    Gemini-2.5-Flash~\citep{gemini1.5} & 63.2 & 5.2 & 57.6 & 4.8 & 69.0 & 6.0 & 5.3 \\
    Gemini-3-Pro-Preview~\citep{gemini1.5} & 82.1 & \textbf{9.0} & 68.7 & \underline{6.7} & 83.6 & \underline{8.5} & \underline{8.1} \\
    GPT-5.1~\citep{gpt5} & 71.2 & 5.7 & 64.8 & 4.8 & 73.0 & 5.8 & 5.4 \\
    
    \cmidrule{1-8}
    \skyblue\multicolumn{8}{c}{\textit{Open-source LLMs}}\\
    \cmidrule{1-8}
    DeepSeek-VL-7B~\citep{deepseek-vl} & 9.7 & 0.4 & 5.9 & 0.3 & 31.4 & 1.4 & 0.7 \\
    Intern-VL-3.5-241B ~\citep{internvl3.5} & \textbf{54.3} & \textbf{3.5} & \textbf{49.2} & \textbf{2.5} & \underline{60.2} & \underline{4.3} & \underline{3.4} \\
    Intern-VL-3.5-30B ~\citep{internvl3.5} & 20.3 & 0.8 & 15.3 & 0.7 & 41.6 & 2.0 & 1.2 \\
    Qwen3-VL-235B ~\citep{qwen3-vl} & \underline{49.2} & \underline{3.3} & \underline{38.2} & \underline{2.4} & \textbf{65.1} & \textbf{5.1} & \textbf{3.6} \\
    Qwen3-VL-30B ~\citep{qwen3-vl} & 17.3 & 0.8 & 19.0 & 0.9 & 45.2 & 2.2 & 1.3 \\
    GLM-4.5V-106B ~\citep{glm4.1v} & 38.4 & 2.8 & 39.1 & 2.1 & 44.3 & 2.0 & 2.3 \\
    GLM-4.1V-9B ~\citep{glm4.1v} & 12.8 & 0.7 & 16.2 & 0.9 & 39.9 & 1.8 & 1.1 \\
    MiMo-VL-7B-RL ~\citep{mimo-vl} & 11.0 & 0.4 & 10.7 & 0.5 & 34.8 & 1.5 & 0.8 \\
    ChartCoder~\citep{chartcoder} & 48.0 & \textbf{3.5} & 40.5 & 2.3 & 54.7 & 3.9 & 3.2 \\

    \bottomrule
    \end{tabular}
\end{adjustbox}
\caption{\textbf{Quantitative results on RealChart2Code 14 primary LLMs across three tasks: Chart Replication, Chart Reproduction, and Chart Refinement.} For each task, we report Pass Rate (\%) and Score. AVG score is the average of scores across all three tasks. Within each model category, the best score is \textbf{bolded} and the second-best is \underline{underlined}.}
\label{tab:main-results}
\end{table*}

\subsection{Evaluation Metrics}
Our evaluation first assesses functional correctness, which we measure as the Pass Rate: the percentage of generated code snippets that execute successfully in our sandbox environment without errors. Submissions that fail this check are automatically assigned a score of zero.
For all valid outputs, we deploy a multi-agent judging panel that uses a voting system to score visual accuracy. Each chart is assessed on a 3-point scale (0, 1, or 2) across eight key criteria: chart type, spatial layout, text elements, axis configuration, color scheme, style, component completeness, and data pattern consistency. Notably, for the Chart Reproduction task, Data Pattern Consistency is evaluated programmatically. We perform a code-level comparison to ensure the model's data handling is identical to the reference implementation, rather than relying on visual inspection. Beyond these core accuracy metrics, our evaluation also includes a qualitative assessment of the chart's design, scored on Visual Clarity, Compositional Balance, and Typographic Quality. The complete scoring rubrics and prompts used for our automated evaluation are detailed in Appendix~\ref{app-sub:metrics}.

\section{Experiments}

\subsection{Experiments Setup}
\paragraph{Baseline Models}
We evaluate 14 widely-used proprietary and open-source Large Language Models. For proprietary models, we evaluate five leading models: Anthropic's flagship models, Claude-4.5-Sonnet and Claude-4.5-Opus~\citep{claude}; OpenAI's advanced model, GPT-5.1~\citep{gpt5}; and Google's Gemini 3 Pro Preview and Gemini 2.5 Flash~\citep{gemini1.5}. For open-weight models, we select 9 competitive models with parameter sizes ranging from 7B to 241B.

\paragraph{Benchmark Setup}
To ensure a comprehensive evaluation, in addition to our primary \texttt{RealChart2Code} benchmark, we also evaluated model performance on two established chart-to-code benchmarks: Plot2Code~\citep{plot2code} and ChartMimic~\citep{chartmimic}.

\begin{table*}[!t]

\small
\centering

\begin{adjustbox}{width=\linewidth}
    \begin{tabular}{l|cc|cc|ccc}
    \toprule
    \multirow{2}*{\makecell*[l]{Model}} 
    & \multicolumn{4}{c|}{\makecell*[c]{ChartMimic}}
    & \multicolumn{3}{c}{\makecell*[c]{Plot2Code}} \\
    & \multicolumn{2}{c|}{Direct Mimic} & \multicolumn{2}{c|}{Customized Mimic} & & & \\
    \cmidrule(lr){2-3} \cmidrule(lr){4-5}
    & Pass (\%) & Score & Pass (\%) & Score & Pass (\%) & Text & Rating \\
    \midrule
    
    \skyblue\multicolumn{8}{c}{\textit{Closed-source LLMs}}\\
    \cmidrule{1-8}
    Claude-4.5-Sonnet ~\citep{claude} & \textbf{100.0} & 90.1 & \underline{99.5} & 91.5 & 93.9 & 73.3 & 8.8 \\
    Claude-4.5-Opus ~\citep{claude} & \underline{98.5} & \underline{92.3} & \textbf{100.0} & \underline{95.1} & \textbf{99.6} & 75.4 & \underline{9.0} \\
    Gemini-2.5-Flash~\citep{gemini1.5} & 88.5 & 69.8 & 89.1 & 74.4 & 87.9 & 72.8 & 8.6 \\
    Gemini-3-Pro-Preview~\citep{gemini1.5} & 97.3 & \textbf{96.0} & \textbf{100.0} & \textbf{96.8} & 90.9 & \underline{79.5} & \textbf{9.6} \\
    GPT-5.1~\citep{gpt5} & 97.8 & 89.8 & 98.5 & 91.2 & \underline{98.5} & \textbf{82.6} & 8.8 \\

    \cmidrule{1-8}
    \skyblue\multicolumn{8}{c}{\textit{Open-source LLMs}}\\
    \cmidrule{1-8}
    DeepSeek-VL-7B~\citep{deepseek-vl} & 41.3 & 19.7 & 59.3 & 37.6 & 64.4 & 32.6 & 2.3 \\
    Intern-VL-3.5-241B ~\citep{internvl3.5} & 90.5 & \underline{76.9} & 90.7 & \underline{78.4} & \textbf{90.2} & \underline{69.0} & 7.6 \\
    Intern-VL-3.5-30B ~\citep{internvl3.5} & 85.6 & 68.1 & 87.2 & 68.5 & 88.6 & 66.3 & 6.7 \\
    Qwen3-VL-235B ~\citep{qwen3-vl} & \textbf{91.5} & \textbf{80.4} & \textbf{92.7} & \textbf{81.8} & \underline{89.8} & \textbf{69.8} & \textbf{8.9} \\
    Qwen3-VL-30B ~\citep{qwen3-vl} & 89.5 & 68.6 & 89.1 & 69.9 & 88.9 & 65.0 & \underline{7.9} \\
    GLM-4.5V-106B ~\citep{glm4.1v} & 87.4 & 72.8 & 86.7 & 71.2 & 84.1 & 61.7 & 5.8 \\
    GLM-4.1V-9B ~\citep{glm4.1v} & 86.6 & 70.2 & 87.1 & 70.1 & 74.1 & 59.1 & 4.9 \\
    MiMo-VL-7B-RL ~\citep{mimo-vl} & 83.2 & 58.5 & 87.7 & 64.7 & 73.5 & 55.5 & 4.3 \\
    ChartCoder~\citep{chartcoder} & \underline{90.9} & 75.3 & \underline{92.1} & 76.7 & 87.9 & 54.5 & 4.7 \\
    
    \bottomrule
    \end{tabular}
\end{adjustbox}
\caption{Quantitative results for 14 LLMs across two benchmarks, ChartMimic and Plot2Code.}
\label{tab:other-results}
\vspace{-1em}
\end{table*}

\paragraph{Evaluation Details}
All experiments were conducted using the standard OpenAI API format, with a chat structure compliant with ChatML~\citep{chatml}. We employed a greedy decoding strategy and set the maximum output token limit to 32,768. If a model did not support this context length, its own maximum limit was used instead. The final reported results are an average of three independent runs. Proprietary models were queried via their official APIs, while open-weight models were served using the SGLang framework. Additional details on prompts and evaluation are provided in Appendix~\ref{app-sub:evaluation_prompt}.

\begin{figure*}[tp]
\centering
\includegraphics[width=\linewidth]{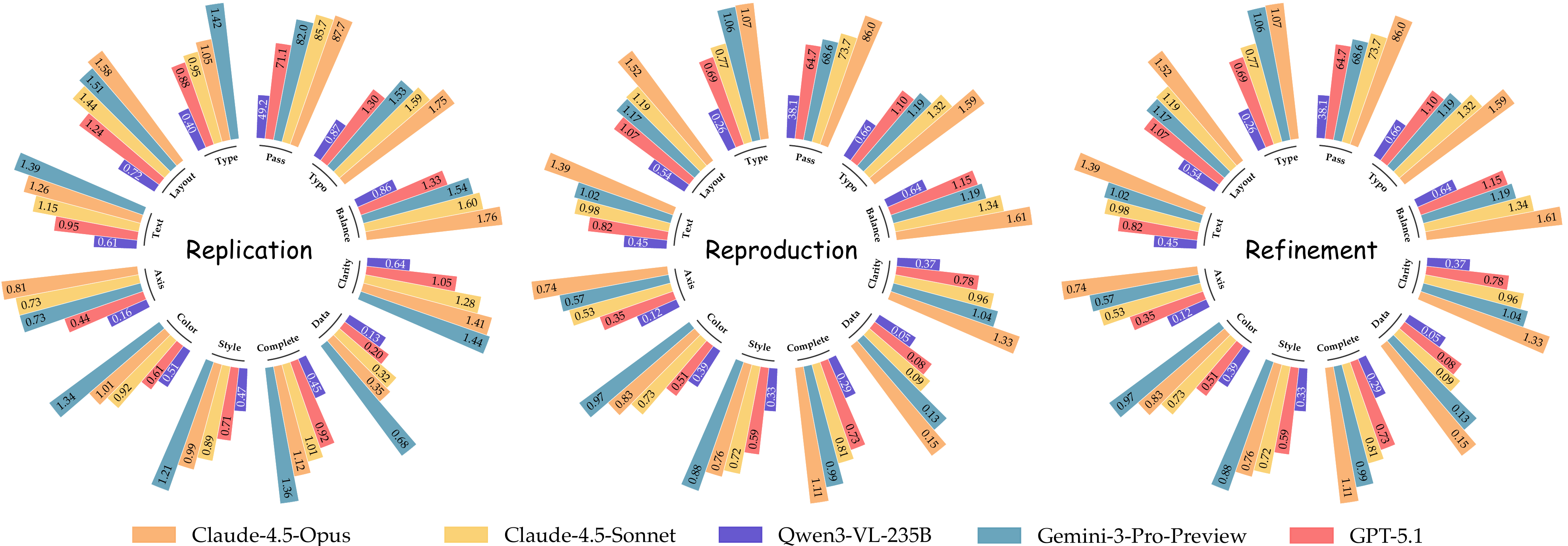}
\caption{Performance breakdown across tasks and metrics for top models. Each radial chart shows scores for eight visual accuracy metrics (Type: chart type consistency, Layout: spatial layout, Text: text elements, Axis: axis configuration, Color: color scheme, Style: style and format, Complete: component completeness, Data: data alignment), execution pass rate (Pass), and three quality metrics (Clarity: visual clarity, Balance: compositional balance, Typo: typographic quality).}
\label{fig:sub_metrics}
\end{figure*}

\subsection{Main Results}
This section presents the evaluation results for 14 leading LLMs on our \texttt{RealChart2Code} benchmark, as well as on the existing ChartMimic and Plot2Code benchmarks. The primary performance on \texttt{RealChart2Code} is detailed in Table~\ref{tab:main-results}, while results on the other benchmarks are shown in Table~\ref{tab:other-results}. For a granular analysis, Figure~\ref{fig:sub_metrics} breaks down the scores by task and sub-metrics, including three quality assessment criteria. The key findings are as follows.

\paragraph{\ding{182} Claude-4.5-Opus leads proprietary models, but a significant performance gap exists between proprietary and open-source models.}
Among proprietary models, Claude-4.5-Opus achieves the highest overall score of 8.2, demonstrating strong and consistent performance across all three tasks. Gemini-3-Pro-Preview follows closely with a score of 8.1, even achieving the top score on the fundamental Chart Replication task at 9.0. In the open-source category, performance is considerably lower. The top-performing models are Qwen3-VL-235B and Intern-VL-3.5-241B, with scores of 3.6 and 3.4, respectively, which are less than half of those from the leading proprietary models. This highlights a significant capability gap on the complex, real-world tasks featured in \texttt{RealChart2Code}.

\paragraph{\ding{183} Strong performance on simpler benchmarks does not guarantee success on \texttt{RealChart2Code}.}
We observe a significant performance disparity for models across different benchmarks, indicating that \texttt{RealChart2Code} provides a much stronger differentiation of model capabilities. For instance, models like Qwen3-VL-235B and Intern-VL-3.5-241B achieve excellent scores above 75 on ChartMimic but experience a drastic performance degradation on \texttt{RealChart2Code}, where their scores drop to 3.6 and 3.4, respectively. This suggests that while previous benchmarks can identify basic competency, their lack of complexity fails to distinguish between models with truly advanced visual reasoning and code generation abilities. In contrast, our benchmark effectively separates the performance of even top-tier models, with Claude-4.5-Opus scoring 8.2 and GPT-5.1 scoring 5.4. Additionally, we find that different models exhibit specific error patterns, which are analyzed in detail in Section~\ref{sec:discussion}.

\section{Discussion}


\begin{table}[t]
    \centering
    \small
    \begin{tabular}{lcc}
        \toprule
        \textbf{Metric} & \textbf{Fleiss' $\kappa$} & \textbf{Cohen's $\kappa$} \\
        & \textbf{(Inter-Agent)} & \textbf{(Agent-Human)} \\
        \midrule
        Type          & 0.921 & 0.91 \\
        Layout        & 0.998 & 0.99 \\
        Text          & 0.896 & 0.79 \\
        Axis          & 0.892 & 0.86 \\
        Color         & 0.879 & 0.72 \\
        Style         & 0.829 & 0.78 \\
        Data          & 0.813 & 0.82 \\
        Completeness  & 0.781 & 0.74 \\
        \midrule
        Average       & 0.824 & 0.83 \\
        \bottomrule
    \end{tabular}
    \caption{Reliability analysis: Fleiss' $\kappa$ for inter-agent agreement and Cohen's $\kappa$ for agreement between multi-agent judge and human evaluations across eight metrics.}
    \label{tab:reliability-analysis}
\end{table}

\begin{figure*}[tp]
\centering
\includegraphics[width=0.96\linewidth]{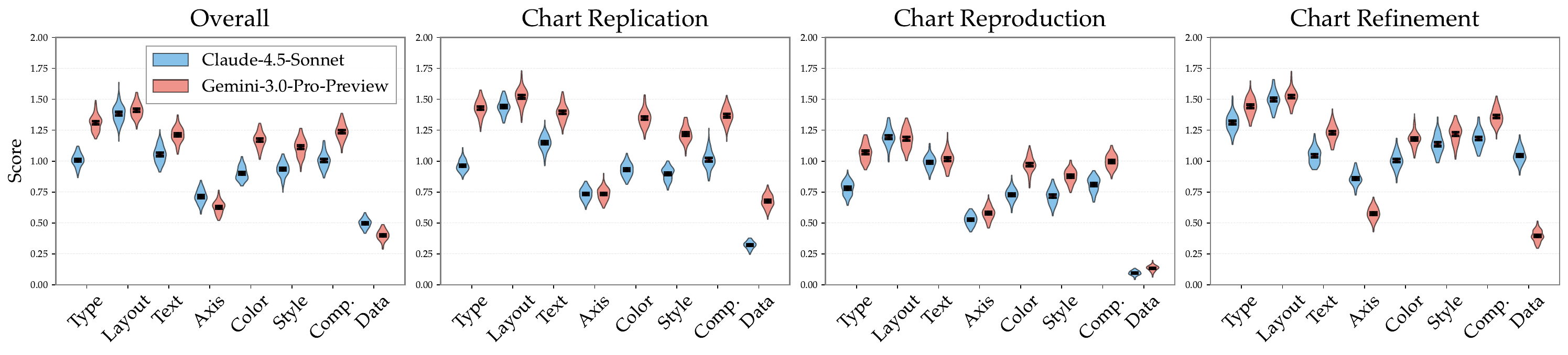}
\caption{Score distributions with 95\% confidence intervals across evaluation metrics for Claude-4.5-Sonnet and Gemini-3.0-Pro-Preview on chart replication, reproduction, and refinement tasks.}
\label{fig:ci}
\vspace{-1em}
\end{figure*}

\subsection{Reliability Analysis}
\label{correlation_analysis}
We validate the robustness of our automated evaluation framework through internal consistency and alignment with human judgment (Table \ref{tab:reliability-analysis}, Figure \ref{fig:ci}).
To assess the consensus among agents, we calculated Fleiss' $\kappa$ across the entire \texttt{RealChart2Code} benchmark. The resulting average Inter-Agent $\kappa$ score of 0.8239 indicates that the multi-agent framework maintains high stability throughout the evaluation process.

We further examined the correlation between our judge and human experts using 600 tasks sampled from Claude-4.5-Sonnet results. By computing Cohen's $\kappa$, we observed an average agreement score of 0.83. This strong correlation demonstrates that our automated multi-agent judge effectively captures human preference.
This reliability is visually corroborated by Figure \ref{fig:ci}, which presents the score distributions and 95\% confidence intervals. The distinct distributions combined with narrow intervals confirm that the judge provides a discriminatory and precise assessment of visualization quality.

\begin{figure}[tp]
\centering
\includegraphics[width=0.95\linewidth]{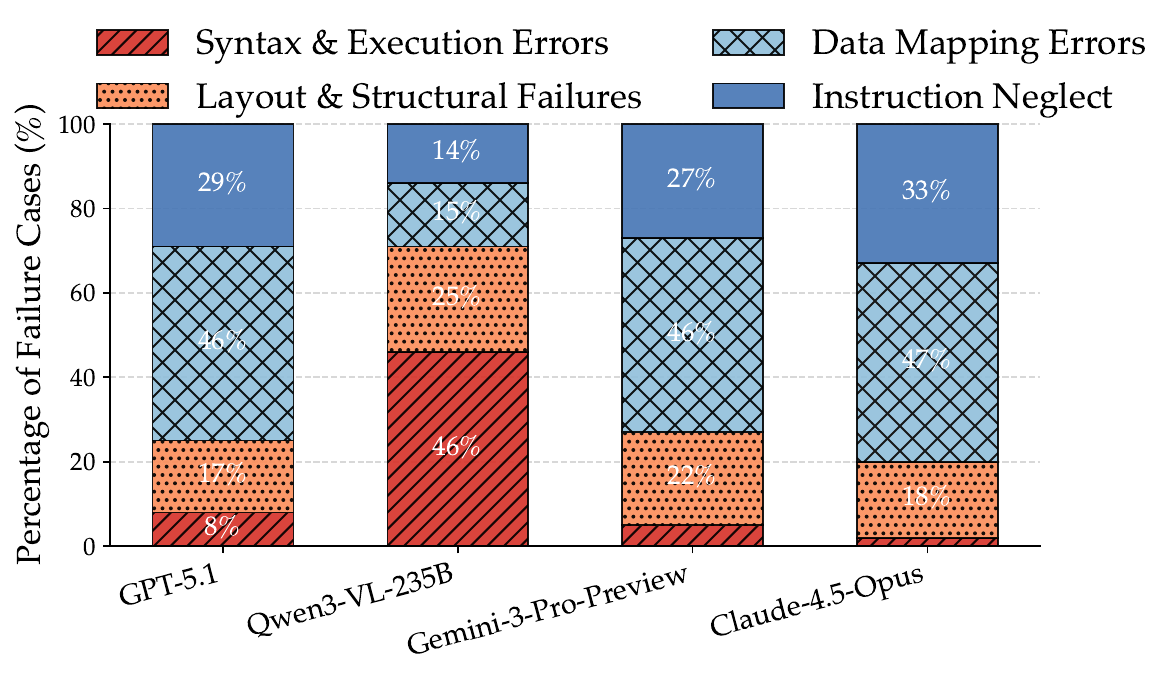}
\caption{Error type distribution across models. Percentages indicate the proportion of each error category in failure cases.}
\label{fig:error_distribution}
\vspace{-1em}
\end{figure}

\subsection{Error Analysis}
\label{sec:discussion}
To gain deeper insights into the limitations of current LLMs, we conducted a comprehensive error analysis covering all failure instances in the \texttt{RealChart2Code} benchmark. This analysis aims to reveal systematic weaknesses in chart-to-code generation and to better understand where existing models break down in practice. We categorize the observed errors into four primary types: (1) Syntax and Execution Errors, (2) Layout and Structural Failures, (3) Data Mapping Errors, and (4) Instruction Neglect, which together capture the major sources of performance degradation.

\paragraph{Divergent Failure Patterns.} We observed a significant disparity in the types of errors exhibited by proprietary versus open-weight models, as illustrated in Figure~\ref{fig:error_distribution}:
\textbf{\ding{182} Open-weight models (e.g., Qwen3-VL, InternVL)} are particularly susceptible to Syntax and Execution Errors. These models frequently hallucinate non-existent libraries or invoke invalid functions, leading to immediate code execution failures. Furthermore, when the code does execute, they often struggle with spatial reasoning, resulting in Layout and Structural Failures, such as overlapping subplots or incorrect grid definitions.
\textbf{\ding{183} Proprietary models (e.g., Claude-4.5, GPT-5.1)}, in contrast, demonstrate robust coding capabilities with minimal syntax errors. Their failures are predominantly Data Mapping Errors, where the visual structure is correct, but specific data series are mapped to the wrong axes or visual attributes do not match the prompt requirements.

\paragraph{Challenges in Iterative Refinement.} The Chart Refinement task reveals a critical weakness in maintaining context. We identified a frequent error mode termed "Regressive Editing." When users request a specific modification, models often successfully apply the fix but inadvertently introduce new errors in previously correct parts of the code. This phenomenon indicates that even state-of-the-art models struggle to balance local code updates with global consistency during multi-turn conversations. Detailed case studies illustrating these error types and specific failure modes are provided in Appendix \ref{app:case_study}.

\subsection{Performance Analysis Across Benchmarks}
\label{sec:cross_benchmark_analysis}

Figure~\ref{fig:benchmark_comparison} compares normalized model performance (0-100\%) on \texttt{RealChart2Code} against \texttt{ChartMimic} and \texttt{Plot2Code}. We observe that while proprietary models like Gemini-3-Pro-Preview saturate existing benchmarks with scores exceeding 91\%, their performance declines to approximately 50\% on our dataset. This gap widens for open-weight models (e.g., Qwen3-VL), where competitive scores ($\sim$85\%) degrade to below 25\%. These results suggest that capabilities demonstrated in simplified synthetic environments fail to transfer effectively to the complex, data-driven scenarios inherent in real-world visualization tasks.
\section{Conclusion and Future Works}

In this work, we addressed the lack of systematic evaluation for Vision Language Models (VLMs) on realistic and complex data visualization tasks. We introduced RealChart2Code, a large-scale benchmark grounded in authentic datasets that assesses capabilities across chart replication, reproduction from raw data, and iterative refinement. Our comprehensive evaluation of 14 leading models demonstrates that while current LLMs are capable of simple plotting, they suffer significant performance degradation when handling complex multi-panel layouts and real-world data. Specifically, we identified a distinct capability gap: proprietary models demonstrate superior visual reasoning, whereas open-weight models frequently struggle with syntax and spatial logic. In future research, we aim to develop automated pipelines for generating high-quality synthetic data to address the scarcity of complex training examples and improve model generalization on intricate layouts.

\section{Limitations}
However, we acknowledge limitations in scope and evaluation. First, our code implementations are currently confined to Matplotlib. While this scope is specific, the granular and imperative nature of Matplotlib code effectively reflects a model's fundamental visualization understanding and logical reasoning capabilities, serving as a robust proxy for general plotting skills. Second, although our MLLM-based judges show strong correlation with human experts, they may still fail to detect subtle visual artifacts, such as minor element overlaps or precise color nuances. These areas offer promising directions for future research in more fine-grained visual evaluation.

\bibliography{custom}
\clearpage
\appendix

\renewcommand*\footnoterule{} 

\newcommand\blfootnote[1]{%
  \begingroup
  \renewcommand\thefootnote{}\footnote{#1}%
  \addtocounter{footnote}{-1}%
  \endgroup
}

\onecolumn 
\section*{Appendix}
\startcontents[sections]
\printcontents[sections]{l}{1}{\setcounter{tocdepth}{2}}
\twocolumn 

\section{Benchmark Details}

\subsection{Chart Tyeps}
\label{app-sub:chart_type}
RealChart2Code encompasses a diverse collection of 50 chart types to ensure comprehensive coverage of real-world visualization scenarios. Table~\ref{tab:chart_type} presents the frequency distribution of these chart types across all tasks in the benchmark. To avoid inflated counts from task repetition, each unique visualization task is counted only once, as the same task set is used across both single-turn and multi-turn interactions. The distribution reflects the natural occurrence of chart types in practical applications, with fundamental visualization types (e.g., line charts, scatter plots, bar charts) appearing more frequently, while specialized techniques (e.g., parallel coordinates, dendrograms, sankey diagrams) are represented in proportion to their real-world usage. This balanced yet realistic distribution ensures that models are evaluated on both common and advanced visualization capabilities.

\begin{table*}[h]
    \caption{Frequency distribution of 50 chart types in RealChart2Code. Each task is counted once to prevent inflation from multi-turn repetitions. The distribution reflects natural occurrence patterns in real-world visualization applications.}
    \label{tab:chart_type}
    \setlength{\tabcolsep}{3pt}
    \small
    \centering
    \resizebox{\textwidth}{!}{
    \begin{tabular}{lc|lc|lc|lc}
        \toprule
        \textbf{Type} & \textbf{Count} & \textbf{Type} & \textbf{Count} & \textbf{Type} & \textbf{Count} & \textbf{Type} & \textbf{Count} \\
        \midrule
        Line Chart & 845 & Scatter Plot & 782 & Bar Chart & 723 & Time Series Plot & 689 \\
        Area Chart & 521 & Histogram & 487 & Box Plot & 456 & Violin Plot & 398 \\
        Heatmap & 374 & Stacked Bar Chart & 342 & Grouped Bar Chart & 318 & Stacked Area Chart & 295 \\
        Bubble Chart & 267 & Pie Chart & 243 & Error Bar Plot & 231 & Density Plot & 218 \\
        Contour Plot & 195 & Radar Chart & 187 & Waterfall Chart & 176 & Network Graph & 164 \\
        Parallel Coordinates & 152 & Sankey Diagram & 143 & Treemap & 138 & Sunburst Chart & 127 \\
        Ridgeline Plot & 119 & Slope Chart & 112 & Stream Graph & 105 & Dendrogram & 98 \\
        Funnel Chart & 91 & Candlestick Chart & 87 & Gantt Chart & 82 & Chord Diagram & 76 \\
        Alluvial Diagram & 71 & Population Pyramid & 67 & Marimekko Chart & 63 & Lollipop Chart & 58 \\
        Dumbbell Plot & 54 & Pair Plot & 49 & Dot Plot & 45 & Word Cloud & 41 \\
        Waffle Chart & 38 & 2D Histogram & 34 & Hexbin Plot & 31 & Jitter Plot & 27 \\
        Joy Plot & 24 & Autocorrelation Plot & 21 & Ternary Plot & 18 & Circular Barplot & 15 \\
        Polar Chart & 12 & Stripplot & 9 & \multicolumn{2}{c|}{} & \multicolumn{2}{c}{} \\
        \bottomrule
    \end{tabular}}
    \vspace{0.5em}
    \\
\end{table*}

\subsection{Render Envirnment}
\label{app-sub:envirnment}
To ensure the safe, consistent, and reproducible execution of all model-generated code, we constructed a dedicated sandboxed environment containerized using Docker. This environment, based on Python 3.13, provides an isolated and standardized platform pre-installed with a comprehensive suite of libraries for data analysis and visualization. Foundational libraries include Pandas for data manipulation and NumPy for numerical operations. For visualization, the environment is equipped with Matplotlib as the primary plotting library, complemented by specialized libraries such as Seaborn for high-level statistical graphics, Plotly for interactive charts, Squarify for treemaps, and others like scikit-learn and statsmodels to support diverse charting requirements. All evaluations are conducted on a server with 128 CPU cores and 1024 GB of RAM, where each code execution runs within its container without network access and is subject to a 120 seconds timeout. This fully-specified setup eliminates system variability and ensures a fair and secure assessment of each model's capabilities.
\section{Benchmark Data Curation Details} 
\label{app:data_curation}

\subsection{Design Principles} The design of \texttt{RealChart2Code} is driven by the necessity to evaluate Vision Language Models (VLMs) not just on their ability to generate code, but on their capacity to perceive, interpret, and reconstruct complex visual data. Our curation process is guided by four core principles:

\begin{itemize} \item \textbf{Grounded in Authentic Data}: Unlike benchmarks that rely on synthetic or simplified data, \texttt{RealChart2Code} is constructed entirely from real-world datasets sourced from Kaggle. This ensures that models are tested against the noise, scale, and irregularity inherent in actual data science workflows. The visualizations reflect genuine analytical intent, moving beyond arbitrary plotting to meaningful data storytelling.
\item \textbf{Visual-Grounded Reverse Engineering}: Distinct from text-to-visualization tasks where models generate charts from open-ended prompts, our benchmark focuses on \textit{Chart Replication} and \textit{Reproduction}. This requires the model to strictly adhere to a visual reference, testing its ability to translate pixel-level visual information (layout, styling, color mapping) into executable matplotlib code with high fidelity.

\item \textbf{Complexity and Diversity}: We explicitly target the "complexity gap" in current evaluations. The benchmark includes a wide spectrum of chart types (over 50 distinct types) and complex composite layouts (e.g., multi-panel figures, dual-axis plots). This compositional complexity probes the model's spatial reasoning and logical understanding of library constraints.

\item \textbf{Iterative Refinement Capability}: Real-world coding is rarely a one-shot process. By incorporating a dedicated \textit{Chart Refinement} module based on error injection, we assess the model's ability to diagnose issues, understand user feedback, and modify existing code without breaking its functionality—a critical skill for collaborative AI assistants.

\end{itemize}

\subsection{Data Collection and Filtering} The foundation of \texttt{RealChart2Code} is a rigorous data selection pipeline designed to filter massive open-source repositories into a high-quality curation. We utilized Kaggle as our primary source, strictly adhering to scientific research licensing terms. The process was executed in two phases:
\paragraph{Phase 1: Automated Large-Scale Screening} We initiated the process with a pool of over 8,000 candidate datasets, comprising more than 100,000 files and roughly 30 billion data rows. To ensure data quality and relevance, we applied a quantitative filtering mechanism based on community engagement metrics. Datasets were ranked and filtered according to vote counts, download frequency, and official usability ratings. This step eliminated incomplete, poorly documented, or trivial datasets.
\paragraph{Phase 2: Expert Curation} From the pre-screened pool, we conducted a manual review to select datasets suitable for complex visualization tasks. The selection criteria focused on data richness (e.g., sufficient columns for multi-dimensional plotting) and domain diversity. This resulted in a final selection of 1,036 high-quality datasets. The final curated collection comprises 3,271 raw data files with approximately 860 million rows, providing a robust testbed for data-intensive generation tasks.

\subsection{Visualization Task Design} Using the 1,036 curated datasets, we constructed 1,016 unique visualization scenarios. To ensure comprehensive coverage of analytical tasks, the design process was guided by a taxonomy of 7 high-level visualization intents and over 50 distinct chart types.

The taxonomy of intents includes: \begin{itemize} \item \textbf{Correlation}: Tasks involving Scatter Plots, Bubble Charts, Pairwise Plots, and Heatmaps to show relationships between variables. \item \textbf{Deviation}: Visualizations such as Diverging Bar Charts and Area Charts to highlight variations from a baseline. \item \textbf{Ranking}: Ordered Bar Charts, Lollipop Charts, and Slope Charts to display comparative rankings. \item \textbf{Distribution}: Histograms, Box Plots, Violin Plots, and Density Plots to analyze data spread. \item \textbf{Composition}: Stacked Bar Charts, Treemaps, and Pie Charts to show part-to-whole relationships. \item \textbf{Change}: Line Charts, Time Series Decompositions, and Area Charts to visualize trends over time. \item \textbf{Groups}: Cluster Plots and Parallel Coordinates to reveal cluster structures within data. \end{itemize}

These 1,016 visualizations form the basis for the \textit{Chart Replication} (Image $\to$ Code) and \textit{Chart Reproduction} (Image + Data $\to$ Code) tasks, totaling 2,032 instances.
\subsection{Ground-Truth Code Implementation} To establish a high-quality "Gold Standard," we avoided using model-generated code as ground truth. Instead, an in-house team of five expert Python developers manually implemented the solution code for all 1,016 visualizations. The implementation strictly utilizes \texttt{matplotlib} and standard data processing libraries (e.g., \texttt{pandas}, \texttt{numpy}). The code was rigorously tested to ensure it is:
\ding{182} Executable: Runs error-free in the evaluation sandbox.
\ding{183} Reproducible: Accurately regenerates the target visualization from the provided data.
\ding{184} Idiomatic: Follows standard coding practices, ensuring the benchmark evaluates the generation of maintainable and clean code.

\subsection{Error Injection and Refinement Tasks} To construct the \textit{Chart Refinement} subset, we manually introduced errors into a selected subset of the ground-truth implementations. This process generated 864 distinct refinement tasks. The injected errors cover a diverse range of categories to simulate real-world debugging scenarios: \begin{itemize} \item \textbf{Visual/Stylistic Errors}: Overlapping elements, incorrect color schemes, missing legends, or illegible labels. \item \textbf{Data Mapping Errors}: Assigning incorrect columns to axes, wrong aggregation methods, or incorrect data filtering. \item \textbf{Chart Type Errors}: Using a suboptimal or incorrect chart type for the given data type (e.g., using a bar chart for continuous time-series data). \end{itemize} In these tasks, the model is presented with the "flawed" code and the resulting incorrect chart, along with a natural language instruction to correct the issue, requiring multi-turn reasoning and code editing capabilities.

\subsection{Quality Control}
To guarantee the robustness and reliability of \texttt{RealChart2Code}, we implemented a strict multi-stage quality control protocol involving peer review, automated execution checks, and consensus-based adjudication.

\paragraph{Ground-Truth Code Verification}
Given that our benchmark relies on human-authored code as the gold standard, ensuring the correctness and quality of these scripts is paramount. We established a **Cross-Validation Peer Review Workflow**:
\begin{enumerate}
    \item \textbf{Execution Sanity Check}: All ground-truth scripts were first run in our standardized sandbox environment. Scripts that failed to execute, produced warnings, or timed out were automatically flagged for revision.
    \item \textbf{Visual Fidelity Review}: Each successful visualization was inspected by a secondary expert who did not author the code. This reviewer verified the chart against a three-point checklist: (1) \textit{Data Accuracy} (does the chart correctly represent the underlying raw data?), (2) \textit{Visual Clarity} (are labels, legends, and layouts legible and overlap-free?), and (3) \textit{Idiomatic Coding} (does the implementation follow standard Matplotlib best practices?).
    \item \textbf{Adjudication}: Any code flagged as sub-optimal during peer review was returned to the original author for correction. If a disagreement persisted regarding the implementation style, a senior lead made the final decision to ensure consistency across the benchmark.
\end{enumerate}

\paragraph{Multi-turn Refinement Verification}
For the Chart Refinement tasks, ensuring the logical consistency between the intentionally flawed code, the rendered image, and the correction instruction was critical. We employed a **Triple-Verification Strategy** similar to the data curation phase:
\begin{enumerate}
    \item \textbf{Triplet Validation}: Each refinement instance consists of a triplet: (Flawed Code/Image, User Instruction, Corrected Ground Truth). These triplets were reviewed by independent annotators.
    \item \textbf{Logic \& Consistency Check}: Annotators verified two specific conditions: First, \textit{Error Visibility}—confirming that the injected error (e.g., "incorrect axis range") was clearly distinguishable in the rendered image. Second, \textit{Solvability}—ensuring that the user instruction provided sufficient information for a model to deduce the correct fix without ambiguity.
    \item \textbf{Majority Vote \& Consensus}: A task was only accepted if it received a unanimous ``Pass" vote. In cases where the error was deemed too subtle or the instruction too vague (partial agreement), the task underwent a consensus discussion phase to polish the prompt. Triplets failing to reach consensus were discarded to prevent model confusion.
\end{enumerate}
This rigorous verification loop ensures that \texttt{RealChart2Code} provides a fair, noise-free, and logically sound evaluation environment for multimodal code generation.
\section{Evaluation Details}

\subsection{Evaluation Metrics}
\label{app-sub:metrics}

This section details the comprehensive scoring rubric used to evaluate the generated visualizations across all three tasks. Our evaluation framework comprises two primary dimensions: \textbf{Visual Structure Alignment} for assessing fidelity to reference charts, and \textbf{Data Alignment} for verifying computational correctness in code reproduction.

\subsubsection{Chart Replication and Chart Refinement Tasks}

For Chart Replication (Task 1) and Chart Refinement (Task 3), evaluation focuses on visual structure alignment with the reference chart. Each metric uses a 3-level scoring system (0/1/2).

\paragraph{Visual Structure Alignment (3-Level Scoring: 0/1/2)}
Assess whether the generated chart replicates the structural elements of the reference chart with precision.

\begin{enumerate}
\item \textbf{Chart Type Consistency}
\begin{itemize}
    \item \textit{Question:} Does the generated chart use identical chart types as the reference for all visualizations (e.g., line, bar, scatter, pie, heatmap)?
    \item \textit{Scoring Criteria:}
    \begin{description}
        \item[Score 2] All chart types match exactly across all subplots, including primary chart types and any overlaid secondary elements (e.g., dual-axis charts, combination charts).
        \item[Score 1] Primary chart types match for all subplots, but minor secondary elements differ (e.g., missing auxiliary bar overlay on a line chart, absent trendline, missing error bars on primary plot).
        \item[Score 0] Any primary chart type differs (e.g., bar chart used where line chart required), or multiple subplots have type mismatches, or fundamental chart category is wrong (e.g., scatter instead of bar).
    \end{description}
\end{itemize}

\item \textbf{Spatial Layout Consistency}
\begin{itemize}
    \item \textit{Question:} Does the generated chart replicate the reference chart's exact component arrangement (subplot grid dimensions, relative positioning of all visual elements)?
    \item \textit{Scoring Criteria:}
    \begin{description}
        \item[Score 2] Exact grid structure (e.g., 2×3 layout), identical subplot positioning, precise relative placement of all elements (legends, annotations, insets) within ±2\% of reference dimensions.
        \item[Score 1] Correct grid structure and subplot count, but minor positioning deviations (legend slightly displaced, annotation shifted but still in correct subplot, spacing variations within 5-15\% of reference).
        \item[Score 0] Incorrect grid dimensions (e.g., 1×4 instead of 2×2), wrong number of subplots, major element misplacement (legend in wrong subplot, annotations in incorrect panels), or spatial relationships fundamentally altered.
    \end{description}
\end{itemize}

\item \textbf{Text Element Consistency}
\begin{itemize}
    \item \textit{Question:} Does the generated chart contain all textual content from the reference (titles, subtitles, annotations, axis labels) with identical wording, excluding axis tick values?
    \item \textit{Scoring Criteria:}
    \begin{description}
        \item[Score 2] All text content matches exactly in wording, placement, and hierarchy (main title, subplot titles, axis labels, annotations, data labels).
        \item[Score 1] All critical text present (main title, axis labels, subplot titles) with exact wording, but minor discrepancies in secondary elements (1-2 missing annotations, slightly reworded footnotes, minor label omissions that do not affect core interpretation).
        \item[Score 0] Missing or altered critical text (wrong/missing main title, incorrect axis labels, absent subplot titles, 3+ missing annotations), or significant semantic changes to any primary text element.
    \end{description}
\end{itemize}

\item \textbf{Axis Configuration Consistency}
\begin{itemize}
    \item \textit{Question:} Does the generated chart replicate all axis properties from the reference (variable names, units, scale types, ranges, tick intervals) and legend specifications for each subplot?
    \item \textit{Scoring Criteria:}
    \begin{description}
        \item[Score 2] Perfect match for all axis properties (scale type, range, units, label text, tick intervals) and legend specifications (position, order, labels) across all subplots.
        \item[Score 1] Axis variable names and scale types correct, but minor deviations (axis range extends ±10-20\% beyond reference, tick intervals slightly different but reasonable, legend order swapped but all entries present, units abbreviated differently but semantically equivalent).
        \item[Score 0] Incorrect axis scale type (log vs. linear), wrong variable assignments, axis range drastically different (>30\% deviation or missing critical data region), missing axis labels, incorrect units, or legend missing critical entries.
    \end{description}
\end{itemize}

\item \textbf{Color Scheme Consistency}
\begin{itemize}
    \item \textit{Question:} Does the generated chart apply the same color mappings as the reference for all data series, categories, and visual elements?
    \item \textit{Scoring Criteria:}
    \begin{description}
        \item[Score 2] Identical color mappings for all elements (data series, categorical encodings, heatmap scales, background, grid, annotations) with exact hex/RGB matches or perceptually indistinguishable equivalents.
        \item[Score 1] Correct color palette family and mapping logic, but minor shade variations (slightly lighter/darker versions of reference colors, saturation differences within 15\%, 1-2 secondary elements use similar but non-exact colors).
        \item[Score 0] Different color palette applied, incorrect categorical color assignments (e.g., Category A colored red instead of blue), reversed color scale (e.g., heatmap scale inverted), or 3+ elements with significantly different colors.
    \end{description}
\end{itemize}

\item \textbf{Style and Format Consistency}
\begin{itemize}
    \item \textit{Question:} Does the generated chart match the reference chart's stylistic attributes (background color, grid styles, marker shapes, line patterns, font families, border styles)?
    \item \textit{Scoring Criteria:}
    \begin{description}
        \item[Score 2] All stylistic attributes match (grid style [solid/dashed/dotted], marker shapes, line widths, line patterns, font families, background colors, border styles, opacity levels).
        \item[Score 1] Core visual style maintained with minor deviations (grid style slightly different [dashed vs. dotted], marker shapes similar but not exact [circle vs. hexagon], line width within ±1px, font family same category [all sans-serif] but different face).
        \item[Score 0] Multiple style mismatches (wrong grid style + wrong markers + wrong font category), fundamentally different aesthetic (dark theme vs. light theme), or critical style elements missing (no gridlines when reference has them, no borders when required).
    \end{description}
\end{itemize}

\item \textbf{Component Completeness}
\begin{itemize}
    \item \textit{Question:} Does the generated chart include all visual components present in the reference chart?
    \item \textit{Scoring Criteria:}
    \begin{description}
        \item[Score 2] Chart contains all visual components from the reference including all data series, markers, annotations, grid elements, legends, statistical overlays (trendlines, confidence intervals, reference lines), decorative features, and auxiliary elements.
        \item[Score 1] All primary components present (all main data series, essential legends, critical annotations, primary axis elements, key statistical markers) but missing 1-3 minor elements (optional secondary annotations, decorative styling features, non-essential reference lines, auxiliary data labels on individual points).
        \item[Score 0] Missing critical components (1+ major data series absent, essential legend missing, key annotations omitted, primary axis elements incomplete, important statistical overlays missing) that significantly affect information completeness or interpretation accuracy.
    \end{description}
\end{itemize}

\item \textbf{Data Pattern Consistency}
\begin{itemize}
    \item \textit{Question:} Does the generated chart visually replicate the reference chart's data patterns (data point positions, trend shapes, distribution profiles, statistical markers)?
    \item \textit{Scoring Criteria:}
    \begin{description}
        \item[Score 2] Data patterns match reference with >95\% positional accuracy, all trend shapes replicated, distribution profiles visually identical, all statistical markers (means, medians, confidence intervals) present and correct.
        \item[Score 1] Core data trends and patterns recognizable (correlation directions correct, major peaks/troughs present, distribution shapes similar), but minor deviations (5-15\% positional variance, slight smoothing differences, 1-2 missing minor statistical markers, interpolation artifacts in small regions).
        \item[Score 0] Data patterns significantly diverged (>15\% positional errors, trend directions incorrect, distribution shapes fundamentally different, major statistical markers missing/wrong), or substantial data missing/added.
    \end{description}
\end{itemize}
\end{enumerate}

\subsubsection{Chart Reproduction Task}

For Chart Reproduction (Task 2), the \textbf{Data Pattern Consistency} metric is replaced with \textbf{Data Alignment}, which performs code-level verification to ensure computational correctness rather than visual similarity.

\paragraph{Data Alignment (3-Level Scoring: 0/1/2)}
Verify that the generated code produces computationally equivalent data transformations and mappings as the reference implementation.

\begin{enumerate}
\item \textbf{Data Source Matching}
\begin{itemize}
    \item Both codes must load data from the same file(s).
    \item Column selections must access the same fields (even if using different syntax).
    \item \textit{Example:} $\texttt{df['col']} \equiv \texttt{df.col} \equiv\texttt{df.loc[:, 'col']} $
\end{itemize}

\item \textbf{Data Transformation Equivalence}

\begin{itemize}
    \item For each subplot, check if the \textbf{final data arrays} passed to plotting functions are equivalent.
    \item \textit{Accept as equivalent:}
    \begin{itemize}
        \item Different groupby syntax producing same result:\\
        \texttt{df.groupby('A')['B'].sum()} $\equiv$\\
        \texttt{df.groupby('A').\\agg(\{'B':'sum'\})['B']}
        
        \item Different filtering methods:\\
        \texttt{df[df['x'] > 0]} $\equiv$ \texttt{df.query('x > 0')} $\equiv$\\
        \texttt{df.loc[df['x'] > 0]}
        
        \item Reordered operations that do not affect output:\\
        \texttt{df.sort\_values('A')\\.reset\_index(drop=True)} $\equiv$\\
        \texttt{df.reset\_index(drop=True)\\.sort\_values('A')}
        
        \item Different intermediate variable names
        \item Different but equivalent aggregation functions on same data
    \end{itemize}
    
    \item \textit{Reject as different:}
    \begin{itemize}
        \item Using different columns: \texttt{df['A']} vs. \texttt{df['B']}
        \item Different aggregation types: \texttt{.sum()} vs. \texttt{.mean()}
        \item Different filters: \texttt{df[df['x'] > 0]} vs.\\
        \texttt{df[df['x'] > 5]}
        \item Different data subsets for same visual element
        \item Missing or extra data transformations that change values
    \end{itemize}
\end{itemize}

\item \textbf{Visual Element Data Mapping}
\begin{itemize}
    \item For each subplot, verify corresponding visual elements use equivalent data.
    \item \textit{Scoring Criteria:}
    \begin{description}
        \item[Score 2] All data transformations are computationally equivalent, and all visual elements map to identical data arrays.
        \item[Score 1] Core data transformations correct with minor acceptable variations (different but equivalent filtering logic, reordered operations producing same output, 1-2 minor data processing steps differ but final arrays match).
        \item[Score 0] Data transformations produce different results (different columns used, different aggregations applied, filters produce different subsets), or visual elements map to non-equivalent data arrays.
    \end{description}
\end{itemize}
\end{enumerate}

\paragraph{Note:} All other Visual Structure Alignment metrics (Chart Type Consistency, Spatial Layout Consistency, Text Element Consistency, Axis Configuration Consistency, Color Scheme Consistency, Style and Format Consistency, and Component Completeness) apply identically to the Chart Reproduction task as described above.

\subsection{Model Details}

\begin{table*}[ht]
\centering
\scalebox{0.6}{
\renewcommand{\arraystretch}{1.35} 
\begin{tabular}{ll|l|l}
\hline
\toprule
\small \textbf{Model} & \small \textbf{Version/HF Checkpoint} & \small \textbf{Release Time} & \small \textbf{Source} \\
\midrule
\multicolumn{4}{c}{\textbf{Proprietary Multimodal Large Language Models}} \\
\midrule
GPT-5.1 \cite{gpt5} & \texttt{gpt-5-1-2025-11-13} & 2025-11-13 & \url{https://openai.com/zh-Hans-CN/index/gpt-5-1/} \\
Claude 4.5 Sonnet \cite{claude} & \texttt{claude-4-5-sonnet-20250929} & 2025-09-29 & \url{https://www.anthropic.com/news/claude-sonnet-4-5} \\
Claude 4.5 Opus \cite{claude} & \texttt{claude-4-5-Opus-20251101} & 2025-11-01 & \url{https://www.anthropic.com/news/claude-opus-4-5} \\
Gemini-2.5-Flash \cite{gemini2.5} & \texttt{gemini-2-5-flash-20250617} & 2025-06-17 & \url{https://deepmind.google/models/gemini/flash/} \\
Gemini-3-Pro-Preivew \cite{gemini2.5} & \texttt{gemini-3-pro-preview} & 2025-11-18 & \url{https://deepmind.google/models/gemini/pro/} \\
\midrule
\multicolumn{4}{c}{\textbf{Open-Source Multimodal Large Language Models}} \\
\midrule
Qwen3-VL-30B \cite{qwen3-vl} & \texttt{Qwen/Qwen3-VL-30B-A3B-Instruct} & 2025-10-10 & \url{https://huggingface.co/Qwen/Qwen3-VL-30B-A3B-Instruct} \\
Qwen3-VL-235B \cite{qwen3-vl} & \texttt{Qwen/Qwen3-VL-235B-A22B-Instruct} & 2025-10-10 & \url{https://huggingface.co/Qwen/Qwen3-VL-235B-A22B-Instruct} \\
Deepseek-VL-7B \cite{deepseek-vl} & \texttt{deepseek-ai/deepseek-vl-7b-base} & 2024-03-09 & \url{https://huggingface.co/deepseek-ai/deepseek-vl-7b-base} \\
MiMo-VL-7B-RL \cite{mimo-vl} & \texttt{XiaomiMiMo/MiMo-VL-7B-RL-2508} & 2025-08-10 & \url{https://huggingface.co/XiaomiMiMo/MiMo-VL-7B-RL-2508} \\
GLM-4.1V-9B \cite{glm4.1v} & \texttt{zai-org/GLM-4.1V-9B-Thinking} & 2025-07-30 & \url{https://huggingface.co/zai-org/GLM-4.1V-9B-Thinking} \\
GLM-4.5V-106B \cite{glm4.1v} & \texttt{zai-org/GLM-4.5V} & 2025-07-30 & \url{https://huggingface.co/zai-org/GLM-4.5V} \\
Intern-VL 3.5 30B \cite{internvl3.5} & \texttt{OpenGVLab/InternVL3\_5-30B-A3B} & 2025-08-25 & \url{https://huggingface.co/OpenGVLab/InternVL3_5-30B-A3B} \\
Intern-VL 3.5 241B \cite{internvl3.5} & \texttt{OpenGVLab/InternVL3\_5-241B-A28B} & 2025-08-25 & \url{https://huggingface.co/OpenGVLab/InternVL3_5-241B-A28B} \\
\bottomrule
\hline
\end{tabular}
}
\label{tab:run_configurations}
\caption{Model details for all models. The release time and model source information are provided for reference.}

\end{table*}

\section{Evaluation Prompts}
\label{app-sub:evaluation_prompt}

\begin{tcolorbox}[
    title={\textbf{Generation Prompt for Chart Replication}},
    breakable,
    colback=blue!5!white,
    colframe=blue!75!black,
    fonttitle=\bfseries,
    colbacktitle=blue!85!black,
    enhanced,
    drop shadow={black!50!white},
    attach boxed title to top left={xshift=5mm, yshift=-2mm},
    boxed title style={size=small, colframe=blue!85!black}
]
\textbf{Role:} \\
You are an expert Python data visualization engineer specializing in matplotlib and seaborn.

\textbf{Task:} \\
Generate executable Python code that precisely replicates the provided reference chart image with pixel-level fidelity.

\textbf{Requirements:} \\
Your code must exactly replicate:

\begin{enumerate}
    \item \textbf{Chart Structure:}
    \begin{itemize}
        \item Chart types
        \item Subplot layout
    \end{itemize}
    
    \item \textbf{Data Representation:}
    \begin{itemize}
        \item Axis properties (labels, units, ranges, scales, ticks)
        \item Data values and trends
        \item Legends (content, position, style)
    \end{itemize}
    
    \item \textbf{Visual Styling:}
    \begin{itemize}
        \item Colors (all elements)
        \item Typography (fonts, sizes)
        \item Line/marker styles (patterns, shapes, widths)
        \item Grids and backgrounds
    \end{itemize}
    
    \item \textbf{Text Content:}
    \begin{itemize}
        \item Titles, annotations, labels (exact wording and placement)
    \end{itemize}
\end{enumerate}

\textbf{Code Standards:}
\begin{itemize}
    \item Use matplotlib and/or seaborn exclusively
    \item Ensure code is self-contained and executable
    \item Generate synthetic data matching reference characteristics
\end{itemize}

\textbf{Output Format:} \\
Provide only the Python code block:

\begin{verbatim}
```python
Your code here
```
\end{verbatim}
\end{tcolorbox}

\begin{tcolorbox}[
    title={\textbf{Evaluation Prompt for Chart Replication}},
    colback=yellow!5!white,
    colframe=yellow!85!black,
    fonttitle=\bfseries,
    colbacktitle=yellow!85!black,
    enhanced,
    drop shadow={black!20!white},
    attach boxed title to top left={xshift=5mm, yshift=-2mm},
    boxed title style={size=small, colframe=yellow!85!black},
    breakable,
    before upper={\parindent0pt}
]

\textbf{Role and Goal:} \\
You are an expert data visualization quality assessor. Your task is to rigorously evaluate Chart B (generated output) against Chart A (reference chart) based on the original task requirements. The evaluation must assess fidelity in visual structure, data representation, and design execution with strict adherence to standards.

\textbf{Evaluation Categories:}

\noindent\textbf{1. Visual Structure Alignment (3-Level Scoring: 0/1/2)} \\
Assess whether Chart B replicates the structural elements of Chart A with precision.

\begin{enumerate}
    \item \textbf{Chart Type Consistency}
    \begin{itemize}
        \item \textbf{Question}: Does Chart B use identical chart types as Chart A for all visualizations?
        \item \textbf{Scoring Criteria}:
        \begin{itemize}
            \item \textbf{Score 2}: All chart types match exactly
            \item \textbf{Score 1}: Primary chart types match
            \item \textbf{Score 0}: Any primary chart type differs
        \end{itemize}
    \end{itemize}
    
    \item \textbf{Spatial Layout Consistency}
    \begin{itemize}
        \item \textbf{Question}: Does Chart B replicate Chart A's exact component arrangement?
        \item \textbf{Scoring Criteria}:
        \begin{itemize}
            \item \textbf{Score 2}: Exact grid structure and positioning
            \item \textbf{Score 1}: Correct grid structure with minor deviations
            \item \textbf{Score 0}: Incorrect grid dimensions or major misplacement
        \end{itemize}
    \end{itemize}
    
    \item \textbf{Text Element Consistency}
    \begin{itemize}
        \item \textbf{Question}: Does Chart B contain all textual content from Chart A?
        \item \textbf{Scoring Criteria}:
        \begin{itemize}
            \item \textbf{Score 2}: All text content matches exactly
            \item \textbf{Score 1}: Critical text present with exact wording
            \item \textbf{Score 0}: Missing or altered critical text
        \end{itemize}
    \end{itemize}
    
    \item \textbf{Axis Configuration Consistency}
    \begin{itemize}
        \item \textbf{Question}: Does Chart B replicate all axis properties from Chart A?
        \item \textbf{Scoring Criteria}:
        \begin{itemize}
            \item \textbf{Score 2}: Perfect match for all axis properties
            \item \textbf{Score 1}: Axis variable names and scale types correct
            \item \textbf{Score 0}: Incorrect axis scale type or range
        \end{itemize}
    \end{itemize}
    
    \item \textbf{Color Scheme Consistency}
    \begin{itemize}
        \item \textbf{Question}: Does Chart B apply the same color mappings as Chart A?
        \item \textbf{Scoring Criteria}:
        \begin{itemize}
            \item \textbf{Score 2}: Identical color mappings for all elements
            \item \textbf{Score 1}: Correct palette family with minor variations
            \item \textbf{Score 0}: Different color palette applied
        \end{itemize}
    \end{itemize}
    
    \item \textbf{Style and Format Consistency}
    \begin{itemize}
        \item \textbf{Question}: Does Chart B match Chart A's stylistic attributes?
        \item \textbf{Scoring Criteria}:
        \begin{itemize}
            \item \textbf{Score 2}: All stylistic attributes match
            \item \textbf{Score 1}: Core visual style maintained
            \item \textbf{Score 0}: Multiple style mismatches
        \end{itemize}
    \end{itemize}
    
    \item \textbf{Data Pattern Consistency}
    \begin{itemize}
        \item \textbf{Question}: Does Chart B visually replicate Chart A's data patterns?
        \item \textbf{Scoring Criteria}:
        \begin{itemize}
            \item \textbf{Score 2}: Data patterns match with >95\% accuracy
            \item \textbf{Score 1}: Core data trends recognizable
            \item \textbf{Score 0}: Data patterns significantly diverged
        \end{itemize}
    \end{itemize}
    
    \item \textbf{Component Completeness}
    \begin{itemize}
        \item \textbf{Scoring Criteria}:
        \begin{itemize}
            \item \textbf{Score 2}: All visual components present
            \item \textbf{Score 1}: All primary components present
            \item \textbf{Score 0}: Missing critical components
        \end{itemize}
    \end{itemize}
\end{enumerate}

\noindent\textbf{2. Execution Quality (3-Level Scoring: 0/1/2)} \\
Evaluate the technical execution quality of Chart B independently.

\begin{enumerate}
    \item \textbf{Visual Clarity}
    \begin{itemize}
        \item \textbf{Scoring Criteria}:
        \begin{itemize}
            \item \textbf{Score 2}: Zero overlap affecting readability
            \item \textbf{Score 1}: Minor overlap limited to non-critical elements
            \item \textbf{Score 0}: Critical overlap affecting primary elements
        \end{itemize}
    \end{itemize}
    
    \item \textbf{Compositional Balance}
    \begin{itemize}
        \item \textbf{Scoring Criteria}:
        \begin{itemize}
            \item \textbf{Score 2}: Optimal spatial distribution
            \item \textbf{Score 1}: Adequate composition with minor issues
            \item \textbf{Score 0}: Poor composition with disproportionate elements
        \end{itemize}
    \end{itemize}
    
    \item \textbf{Typographic Quality}
    \begin{itemize}
        \item \textbf{Scoring Criteria}:
        \begin{itemize}
            \item \textbf{Score 2}: All text appropriately sized and legible
            \item \textbf{Score 1}: Text mostly appropriate with minor issues
            \item \textbf{Score 0}: Text severely undersized or mispositioned
        \end{itemize}
    \end{itemize}
\end{enumerate}

\noindent\textbf{Output Format:} \\
Your response \textbf{MUST} be a single, valid JSON object with no additional text.

\begin{verbatim}
```json
{
  "visual_structure_alignment": {
    "chart_type_consistency": {
      "score": 2,
      "reason": "Detailed explanation"
    },
    "spatial_layout_consistency": {
      "score": 2,
      "reason": "Detailed explanation"
    },
    "text_element_consistency": {
      "score": 2,
      "reason": "Detailed explanation"
    },
    "axis_configuration_consistency": 
    { "score": 2,
      "reason": "Detailed explanation"
    },
    "color_scheme_consistency": {
      "score": 2,
      "reason": "Detailed explanation"
    },
    "style_and_format_consistency": {
      "score": 2,
      "reason": "Detailed explanation"
    },
    "data_pattern_consistency": {
      "score": 2,
      "reason": "Detailed explanation"
    },
    "component_completeness": {
      "score": 2,
      "reason": "Detailed explanation"
    }
  },
  "execution_quality": {
    "visual_clarity": {
      "score": 2,
      "reason": "Detailed explanation"
    },
    "compositional_balance": {
      "score": 2,
      "reason": "Detailed explanation"
    },
    "typographic_quality": {
      "score": 2,
      "reason": "Detailed explanation"
    }
  },
  "improvement_recommendations": 
  "Priority fixes with explanations"
}
\end{verbatim}

\end{tcolorbox}

\begin{tcolorbox}[
    title={\textbf{Generation Prompt for Chart Reproduction}},
    breakable,
    colback=blue!5!white,
    colframe=blue!75!black,
    fonttitle=\bfseries,
    colbacktitle=blue!85!black,
    enhanced,
    drop shadow={black!20!white},
    attach boxed title to top left={xshift=5mm, yshift=-2mm},
    boxed title style={size=small, colframe=blue!85!black},
    before upper={\parindent0pt}]

\textbf{Role:} \\
You are an expert Python data visualization engineer specializing in matplotlib and seaborn.

\textbf{Task:} \\
Analyze the provided data information and reference chart image to identify the data relationships and patterns being visualized. Generate executable Python code that precisely replicates the chart with pixel-level fidelity using the actual data structure described.

\textbf{Inputs:}

\begin{enumerate}
    \item \textbf{Data Information:} 
    \begin{itemize}
        \item Dataset descriptions including shape, columns, data types, and sample rows
        \item Multiple CSV/XLSX files may be provided with their respective schemas
    \end{itemize}
    
    \item \textbf{Reference Chart:}
    \begin{itemize}
        \item Image showing the target visualization to replicate
    \end{itemize}
\end{enumerate}

\textbf{Analysis Requirements:}

Before generating code, identify:
\begin{itemize}
    \item Which dataset(s) and columns are used in the visualization
    \item Data transformations applied (aggregations, calculations, filtering, etc.)
    \item Relationships between multiple datasets (if applicable)
    \item Time series patterns, trends, or statistical relationships displayed
\end{itemize}

\textbf{Code Requirements:}

Your code must exactly replicate:

\begin{enumerate}
    \item \textbf{Data Loading:}
    \begin{itemize}
        \item \textbf{MANDATORY:} Read data from the provided CSV or XLSX files using pandas (pd.read\_csv() or pd.read\_excel())
        \item Use the exact file paths/names mentioned in the data information
        \item Apply necessary data type conversions (dates, numeric values, categories)
        \item Handle multiple files if multiple datasets are referenced
    \end{itemize}
    
    \item \textbf{Data Mapping:}
    \begin{itemize}
        \item \textbf{CRITICAL:} Ensure each data column is correctly mapped to its corresponding visual element in the chart
        \item X-axis must display the correct column data in the proper order
        \item Y-axis must display the correct column data with accurate values
        \item Multiple series/groups must use the correct data columns with proper labels
        \item Verify that data values in the visualization match the source data numerically and categorically
        \item Apply correct aggregations (sum, mean, count, etc.) to match the reference chart
        \item Ensure data filtering and grouping operations produce the exact subsets shown in the visualization
    \end{itemize}
    
    \item \textbf{Chart Structure:}
    \begin{itemize}
        \item Chart types (line, bar, scatter, etc.)
        \item Subplot layout and arrangement
        \item Figure dimensions and aspect ratios
    \end{itemize}
    
    \item \textbf{Data Representation:}
    \begin{itemize}
        \item Axis properties (labels, units, ranges, scales, ticks)
        \item Data values, trends, and patterns from the input datasets
        \item Legends (content, position, style)
        \item Statistical elements (means, medians, trend lines, etc.)
    \end{itemize}
    
    \item \textbf{Visual Styling:}
    \begin{itemize}
        \item Colors for all elements (lines, markers, backgrounds)
        \item Typography (fonts, sizes, weights)
        \item Line/marker styles (patterns, shapes, widths, transparency)
        \item Grids, spines, and backgrounds
    \end{itemize}
    
    \item \textbf{Text Content:}
    \begin{itemize}
        \item Titles, axis labels, annotations (exact wording and placement)
        \item Value labels or data point annotations
    \end{itemize}
\end{enumerate}

\textbf{Code Standards:}
\begin{itemize}
    \item Use matplotlib and/or seaborn exclusively
    \item Load and process data matching the provided data\_info structure
    \item Include necessary data preprocessing (parsing dates, filtering, calculations)
    \item Ensure code is self-contained and executable
    \item Use the actual column names and data types from the input datasets
    \item \textbf{DO NOT generate synthetic data}
\end{itemize}

\textbf{Output Format:} \\
Provide only the Python code block:

\begin{verbatim}
```python
# Your code here
\end{verbatim}

\end{tcolorbox}

\begin{tcolorbox}[
    title={\textbf{Data Evaluation Prompt for Chart Reproduction}},
    colback=orange!5!white,
    colframe=orange!85!black,
    fonttitle=\bfseries,
    colbacktitle=orange!85!black,
    enhanced,
    drop shadow={black!20!white},
    attach boxed title to top left={xshift=5mm, yshift=-2mm},
    boxed title style={size=small, colframe=orange!85!black},
    breakable,
    before upper={\parindent0pt}
]

\textbf{Task:} \\
Compare two Python visualization codes and determine if they use \textbf{functionally equivalent data} for each corresponding subplot.

\textbf{Evaluation Criteria:}

\vspace{5pt}

\noindent\textbf{1. Data Source Matching}
\begin{itemize}
    \item Both codes must load data from the same file(s)
    \item Column selections must access the same fields (even if using different syntax)
    \item Example: $\texttt{df['col']} \equiv \texttt{df.col} \equiv \texttt{df.loc[:, 'col']}$
\end{itemize}

\vspace{5pt}

\noindent\textbf{2. Data Transformation Equivalence} \\
For each subplot, check if the \textbf{final data arrays} passed to plotting functions are equivalent:

\vspace{2pt}

\noindent\textbf{ACCEPT as equivalent:}
\begin{itemize}
    \item Different groupby syntax producing same result:
    \begin{itemize}
        \item \texttt{df.groupby('A')['B'].sum()} ≡ \texttt{df.groupby('A').agg(\{'B':
        'sum'\})['B']}
    \end{itemize}
    \item Different filtering methods:
    \begin{itemize}
        \item \texttt{df[df['x'] > 0]} ≡ \texttt{df.query('x > 0')} ≡ \texttt{df.loc[df['x'] > 0]}
    \end{itemize}
    \item Reordered operations that don't affect output:
    \begin{itemize}
        \item \texttt{df.sort\_values('A').reset
        \_index(drop=True)} ≡ \texttt{df.reset\_index(drop=True).
        sort\_values('A')}
    \end{itemize}
    \item Different intermediate variable names
    \item Different but equivalent aggregation functions on same data
\end{itemize}

\vspace{2pt}

\noindent\textbf{REJECT as different:}
\begin{itemize}
    \item Using different columns: \texttt{df['A']} vs \texttt{df['B']}
    \item Different aggregation types: \texttt{.sum()} vs \texttt{.mean()}
    \item Different filters: \texttt{df[df['x'] > 0]} vs \texttt{df[df['x'] > 5]}
    \item Different data subsets for same visual element
    \item Missing or extra data transformations that change values
\end{itemize}

\vspace{5pt}

\noindent\textbf{3. Visual Element Data Mapping} \\
For each subplot, verify corresponding visual elements use equivalent data:

\vspace{2pt}

\noindent\textbf{Must match:}
\begin{itemize}
    \item X-axis data source and values
    \item Y-axis data source and values  
    \item Color/hue grouping variables
    \item Size/marker data sources
    \item Facet/subplot grouping variables
    \item Filter conditions (if applied to specific elements)
\end{itemize}

\vspace{2pt}

\noindent\textbf{May differ:}
\begin{itemize}
    \item Variable names (\texttt{x\_data} vs \texttt{x\_vals})
    \item Intermediate calculation steps (as long as final result is same)
    \item Code organization (functions vs inline)
\end{itemize}

\vspace{5pt}

\textbf{Verification Steps:}

\begin{enumerate}
    \item \textbf{Identify corresponding subplots} between two codes (by position or title)
    \item \textbf{For each subplot pair}, trace data flow:
    \begin{itemize}
        \item What columns are loaded?
        \item What transformations applied? (group, filter, aggregate, calculate)
        \item What data arrays enter \texttt{plt.plot()}, \texttt{plt.bar()}, \texttt{sns.lineplot()}, etc.?
    \end{itemize}
    \item \textbf{Compare final data semantics}, not code syntax
    \item \textbf{Check ALL subplots} - if any subplot uses different data, return NO
\end{enumerate}

\vspace{5pt}

\textbf{Edge Cases:}

\begin{itemize}
    \item \textbf{Hardcoded values}: If one code uses \texttt{plt.bar([1,2,3], [4,5,6])} and another loads same values from CSV, consider \textbf{EQUIVALENT}
    \item \textbf{Derived columns}: \texttt{df['ratio'] = df['A']/df['B']} must use same columns in both codes
    \item \textbf{Statistical calculations}: \texttt{df['A']
    .mean()} vs \texttt{np.mean(df['
    A'])}
    are \textbf{EQUIVALENT}
\end{itemize}

\vspace{5pt}

\textbf{Output Format:}

\begin{verbatim}
<reasoning>
[Brief analysis of each subplot's data
usage comparison]
</reasoning>

<result>YES</result>  <!-- If all sub-
plots use equivalent data -->
or
<result>NO</result>   <!-- If ANY sub-
plot uses different data -->
\end{verbatim}

\vspace{5pt}

\textbf{Examples:}

\noindent\textbf{Example 1: Should return YES}
\begin{verbatim}
# Code A
data = pd.read_csv('data.csv')
train = data[data['split'] == 'Train']
plt.bar(train['category'],
train['count'])

# Code B  
df = pd.read_csv('data.csv')
filtered = df.query("split == 'Train'")
plt.bar(filtered['category'].values, 
filtered['count'].values)
\end{verbatim}

\end{tcolorbox}

\begin{tcolorbox}[
    title={\textbf{Evaluation Prompt for Chart Reproduction}},
    colback=purple!5!white,
    colframe=purple!85!black,
    fonttitle=\bfseries,
    colbacktitle=purple!85!black,
    enhanced,
    drop shadow={black!20!white},
    attach boxed title to top left={xshift=5mm, yshift=-2mm},
    boxed title style={size=small, colframe=purple!85!black},
    breakable,
    before upper={\parindent0pt}
]

\textbf{Role and Goal:} \\
You are an expert data visualization quality assessor. Your task is to rigorously evaluate Chart B (generated output) against Chart A (reference chart) based on the original task requirements. The evaluation must assess fidelity in visual structure, data representation, and design execution with strict adherence to standards.

\vspace{5pt}

\textbf{Evaluation Categories:}

\vspace{5pt}

\noindent\textbf{1. Visual Structure Alignment (3-Level Scoring: 0/1/2)} \\
Assess whether Chart B replicates the structural elements of Chart A with precision.

\begin{enumerate}
    \item \textbf{Chart Type Consistency}
    \begin{itemize}
        \item \textbf{Question}: Does Chart B use identical chart types as Chart A for all visualizations?
        \item \textbf{Scoring Criteria}:
        \begin{itemize}
            \item \textbf{Score 2}: All chart types match exactly across all subplots
            \item \textbf{Score 1}: Primary chart types match for all subplots
            \item \textbf{Score 0}: Any primary chart type differs
        \end{itemize}
    \end{itemize}
    
    \item \textbf{Spatial Layout Consistency}
    \begin{itemize}
        \item \textbf{Question}: Does Chart B replicate Chart A's exact component arrangement?
        \item \textbf{Scoring Criteria}:
        \begin{itemize}
            \item \textbf{Score 2}: Exact grid structure, identical subplot positioning
            \item \textbf{Score 1}: Correct grid structure and subplot count, but minor positioning deviations
            \item \textbf{Score 0}: Incorrect grid dimensions, wrong number of subplots, major element misplacement
        \end{itemize}
    \end{itemize}
    
    \item \textbf{Text Element Consistency}
    \begin{itemize}
        \item \textbf{Question}: Does Chart B contain all textual content from Chart A?
        \item \textbf{Scoring Criteria}:
        \begin{itemize}
            \item \textbf{Score 2}: All text content matches exactly in wording, placement, and hierarchy
            \item \textbf{Score 1}: All critical text present with exact wording, but minor discrepancies in secondary elements
            \item \textbf{Score 0}: Missing or altered critical text
        \end{itemize}
    \end{itemize}
    
    \item \textbf{Axis Configuration Consistency}
    \begin{itemize}
        \item \textbf{Question}: Does Chart B replicate all axis properties from Chart A?
        \item \textbf{Scoring Criteria}:
        \begin{itemize}
            \item \textbf{Score 2}: Perfect match for all axis properties and legend specifications
            \item \textbf{Score 1}: Axis variable names and scale types correct, but minor deviations
            \item \textbf{Score 0}: Incorrect axis scale type, wrong variable assignments, axis range drastically different
        \end{itemize}
    \end{itemize}
    
    \item \textbf{Color Scheme Consistency}
    \begin{itemize}
        \item \textbf{Question}: Does Chart B apply the same color mappings as Chart A?
        \item \textbf{Scoring Criteria}:
        \begin{itemize}
            \item \textbf{Score 2}: Identical color mappings for all elements with exact hex/RGB matches
            \item \textbf{Score 1}: Correct color palette family and mapping logic, but minor shade variations
            \item \textbf{Score 0}: Different color palette applied, incorrect categorical color assignments
        \end{itemize}
    \end{itemize}
    
    \item \textbf{Style and Format Consistency}
    \begin{itemize}
        \item \textbf{Question}: Does Chart B match Chart A's stylistic attributes?
        \item \textbf{Scoring Criteria}:
        \begin{itemize}
            \item \textbf{Score 2}: All stylistic attributes match
            \item \textbf{Score 1}: Core visual style maintained with minor deviations
            \item \textbf{Score 0}: Multiple style mismatches or fundamentally different aesthetic
        \end{itemize}
    \end{itemize}
    
    \item \textbf{Component Completeness}
    \begin{itemize}
        \item \textbf{Scoring Criteria}:
        \begin{itemize}
            \item \textbf{Score 2}: Chart B contains all visual components from Chart A
            \item \textbf{Score 1}: All primary components present but missing 1-3 minor elements
            \item \textbf{Score 0}: Missing critical components that significantly affect information completeness
        \end{itemize}
    \end{itemize}
\end{enumerate}

\vspace{5pt}

\noindent\textbf{2. Execution Quality (3-Level Scoring: 0/1/2)} \\
Evaluate the technical execution quality of Chart B independently.

\begin{enumerate}
    \item \textbf{Visual Clarity}
    \begin{itemize}
        \item \textbf{Scoring Criteria}:
        \begin{itemize}
            \item \textbf{Score 2}: Zero overlap or occlusion affecting readability
            \item \textbf{Score 1}: Minor overlap limited to 1-2 non-critical elements
            \item \textbf{Score 0}: Critical overlap affecting primary elements that severely impairs interpretation
        \end{itemize}
    \end{itemize}
    
    \item \textbf{Compositional Balance}
    \begin{itemize}
        \item \textbf{Scoring Criteria}:
        \begin{itemize}
            \item \textbf{Score 2}: Optimal spatial distribution with proportional element sizing
            \item \textbf{Score 1}: Adequate composition with acceptable proportions and spacing
            \item \textbf{Score 0}: Poor composition with disproportionate elements
        \end{itemize}
    \end{itemize}
    
    \item \textbf{Typographic Quality}
    \begin{itemize}
        \item \textbf{Scoring Criteria}:
        \begin{itemize}
            \item \textbf{Score 2}: All text appropriately sized relative to chart dimensions
            \item \textbf{Score 1}: Text mostly appropriate with minor issues
            \item \textbf{Score 0}: Text severely undersized or oversized, illegible elements
        \end{itemize}
    \end{itemize}
\end{enumerate}

\vspace{10pt}

\noindent\textbf{Output Format:} \\
Your response \textbf{MUST} be a single, valid JSON object with no additional text. Use this exact structure:

\begin{verbatim}
{
  "visual_structure_alignment": {
    "chart_type_consistency": {
      "score": 1,
      "reason": "Detailed explanation"
    },
    "spatial_layout_consistency": {
      "score": 2,
      "reason": "Detailed explanation"
    },
    "text_element_consistency": {
      "score": 1,
      "reason": "Detailed explanation"
    },
    "axis_configuration_consistency": 
    { "score": 1,
      "reason": "Detailed explanation"
    },
    "color_scheme_consistency": {
      "score": 2,
      "reason": "Detailed explanation"
    },
    "style_and_format_consistency": {
      "score": 1,
      "reason": "Detailed explanation"
    },
    "data_pattern_consistency": {
      "score": 2,
      "reason": "Detailed explanation"
    },
    "component_completeness": {
      "score": 1,
      "reason": "Detailed explanation"
    }
  },
  "execution_quality": {
    "visual_clarity": {
      "score": 1,
      "reason": "Detailed explanation"
    },
    "compositional_balance": {
      "score": 2,
      "reason": "Detailed explanation"
    },
    "typographic_quality": {
      "score": 2,
      "reason": "Detailed explanation"
    }
  },
  "improvement_recommendations": "
  Priority fixes: (1) Add missing 
  auxiliary bar overlay..."
}
\end{verbatim}

\end{tcolorbox}

\begin{tcolorbox}[
    title={\textbf{Generation Prompt for Chart Refinement}},
    colback=teal!5!white,
    colframe=teal!85!black,
    fonttitle=\bfseries,
    colbacktitle=teal!85!black,
    enhanced,
    drop shadow={black!20!white},
    attach boxed title to top left={xshift=5mm, yshift=-2mm},
    boxed title style={size=small, colframe=teal!85!black},
    breakable,
    before upper={\parindent0pt}
]

\textbf{Role:} \\
You are an expert Python data visualization engineer specializing in matplotlib and seaborn, with strong skills in chart debugging and correction.

\textbf{Task:} \\
You will receive:
\begin{enumerate}
    \item An \textbf{image of a chart} that contains visual errors or issues
    \item An \textbf{instruction} describing what needs to be corrected
\end{enumerate}

Your goal is to generate corrected, executable Python code that fixes the issues identified in the instruction while maintaining other aspects of the original chart.

\textbf{Instruction Input Format:} \\
The instruction will be provided in natural language describing the required corrections, such as:
\begin{itemize}
    \item "Fix the incorrect color scheme - bars should be blue instead of red"
    \item "Correct the y-axis scale which is currently inverted"
    \item "Fix the overlapping labels on the x-axis"
    \item "Correct the legend positioning - it's currently obscuring the data"
    \item "Fix the incorrect data values in the second subplot"
    \item "Adjust the title alignment and font size"
    \item "Fix the misaligned text labels"
    \item "Correct the vertical alignment of annotations"
\end{itemize}

\textbf{Your Responsibilities:}

\begin{enumerate}
    \item \textbf{Analyze the Chart Image:}
    \begin{itemize}
        \item Identify the chart structure, type, and layout
        \item Understand the current state (including the errors)
        \item Note all visual elements and styling
    \end{itemize}
    
    \item \textbf{Interpret the Instruction:}
    \begin{itemize}
        \item Understand what specific corrections are needed
        \item Determine which elements should be modified
        \item Preserve all elements not mentioned in the instruction
    \end{itemize}
    
    \item \textbf{Generate Corrected Code that Must Exactly Replicate:}
    
    \begin{itemize}
        \item \textbf{Chart Structure:}
        \begin{itemize}
            \item Chart types
            \item Subplot layout
        \end{itemize}
        
        \item \textbf{Data Representation:}
        \begin{itemize}
            \item Axis properties (labels, units, ranges, scales, ticks)
            \item Data values and trends
            \item Legends (content, position, style)
        \end{itemize}
        
        \item \textbf{Visual Styling:}
        \begin{itemize}
            \item Colors (all elements)
            \item Typography (fonts, sizes)
            \item Line/marker styles (patterns, shapes, widths)
            \item Grids and backgrounds
        \end{itemize}
        
        \item \textbf{Text Content \& Alignment:}
        \begin{itemize}
            \item Titles, annotations, labels (exact wording and placement)
            \item Text alignment (horizontal and vertical: left/center/right, top/center/bottom)
            \item Text rotation and orientation
            \item Spacing and positioning relative to chart elements
        \end{itemize}
    \end{itemize}
\end{enumerate}

\textbf{Code Standards:}
\begin{itemize}
    \item Use matplotlib and/or seaborn exclusively
    \item Ensure code is self-contained and executable
    \item Generate synthetic data that matches the reference characteristics
    \item Apply corrections precisely as instructed
    \item Maintain all other visual properties of the original chart
    \item Pay special attention to alignment properties (ha, va, align parameters)
\end{itemize}

\textbf{Output Format:} \\
Provide only the corrected Python code block:

\begin{verbatim}
```python
# Your corrected code here
\end{verbatim}

\end{tcolorbox}

\begin{tcolorbox}[
    title={\textbf{Evaluation Prompt for Chart Refinement}},
    colback=purple!5!white,
    colframe=purple!85!black,
    fonttitle=\bfseries,
    colbacktitle=purple!85!black,
    enhanced,
    drop shadow={black!20!white},
    attach boxed title to top left={xshift=5mm, yshift=-2mm},
    boxed title style={size=small, colframe=purple!85!black},
    breakable,
    before upper={\parindent0pt}
]

\textbf{Role and Goal:} \\
You are an expert data visualization quality assessor. Your task is to rigorously evaluate Chart B (generated output) against Chart A (reference chart) based on the original task requirements. The evaluation must assess fidelity in visual structure, data representation, and design execution with strict adherence to standards.

\vspace{10pt}

\textbf{Evaluation Categories:}

\vspace{5pt}

\noindent\textbf{1. Visual Structure Alignment (3-Level Scoring: 0/1/2)} \\
Assess whether Chart B replicates the structural elements of Chart A with precision.

\vspace{5pt}

\begin{enumerate}
    \item \textbf{Chart Type Consistency}
    \begin{itemize}
        \item \textbf{Question}: Does Chart B use identical chart types as Chart A for all visualizations (e.g., line, bar, scatter, pie, heatmap)?
        \item \textbf{Scoring Criteria}:
        \begin{itemize}
            \item \textbf{Score 2}: All chart types match exactly across all subplots, including primary chart types and any overlaid secondary elements (e.g., dual-axis charts, combination charts)
            \item \textbf{Score 1}: Primary chart types match for all subplots, but minor secondary elements differ (e.g., missing auxiliary bar overlay on a line chart, absent trendline, missing error bars on primary plot)
            \item \textbf{Score 0}: Any primary chart type differs (e.g., bar chart used where line chart required), or multiple subplots have type mismatches, or fundamental chart category is wrong (e.g., scatter instead of bar)
        \end{itemize}
    \end{itemize}
    
    \item \textbf{Spatial Layout Consistency}
    \begin{itemize}
        \item \textbf{Question}: Does Chart B replicate Chart A's exact component arrangement (subplot grid dimensions, relative positioning of all visual elements)?
        \item \textbf{Scoring Criteria}:
        \begin{itemize}
            \item \textbf{Score 2}: Exact grid structure (e.g., 2×3 layout), identical subplot positioning, precise relative placement of all elements (legends, annotations, insets) within ±2\% of reference dimensions
            \item \textbf{Score 1}: Correct grid structure and subplot count, but minor positioning deviations (legend slightly displaced, annotation shifted but still in correct subplot, spacing variations within 5-15\% of reference)
            \item \textbf{Score 0}: Incorrect grid dimensions (e.g., 1×4 instead of 2×2), wrong number of subplots, major element misplacement (legend in wrong subplot, annotations in incorrect panels), or spatial relationships fundamentally altered
        \end{itemize}
    \end{itemize}
    
    \item \textbf{Text Element Consistency}
    \begin{itemize}
        \item \textbf{Question}: Does Chart B contain all textual content from Chart A (titles, subtitles, annotations, axis labels) with identical wording, excluding axis tick values?
        \item \textbf{Scoring Criteria}:
        \begin{itemize}
            \item \textbf{Score 2}: All text content matches exactly in wording, placement, and hierarchy (main title, subplot titles, axis labels, annotations, data labels)
            \item \textbf{Score 1}: All critical text present (main title, axis labels, subplot titles) with exact wording, but minor discrepancies in secondary elements (1-2 missing annotations, slightly reworded footnotes, minor label omissions that don't affect core interpretation)
            \item \textbf{Score 0}: Missing or altered critical text (wrong/missing main title, incorrect axis labels, absent subplot titles, 3+ missing annotations), or significant semantic changes to any primary text element
        \end{itemize}
    \end{itemize}
    
    \item \textbf{Axis Configuration Consistency}
    \begin{itemize}
        \item \textbf{Question}: Does Chart B replicate all axis properties from Chart A (variable names, units, scale types, ranges, tick intervals) and legend specifications for each subplot?
        \item \textbf{Scoring Criteria}:
        \begin{itemize}
            \item \textbf{Score 2}: Perfect match for all axis properties (scale type, range, units, label text, tick intervals) and legend specifications (position, order, labels) across all subplots
            \item \textbf{Score 1}: Axis variable names and scale types correct, but minor deviations (axis range extends ±10-20\% beyond reference, tick intervals slightly different but reasonable, legend order swapped but all entries present, units abbreviated differently but semantically equivalent)
            \item \textbf{Score 0}: Incorrect axis scale type (log vs linear), wrong variable assignments, axis range drastically different (>30\% deviation or missing critical data region), missing axis labels, incorrect units, or legend missing critical entries
        \end{itemize}
    \end{itemize}
    
    \item \textbf{Color Scheme Consistency}
    \begin{itemize}
        \item \textbf{Question}: Does Chart B apply the same color mappings as Chart A for all data series, categories, and visual elements?
        \item \textbf{Scoring Criteria}:
        \begin{itemize}
            \item \textbf{Score 2}: Identical color mappings for all elements (data series, categorical encodings, heatmap scales, background, grid, annotations) with exact hex/RGB matches or perceptually indistinguishable equivalents
            \item \textbf{Score 1}: Correct color palette family and mapping logic, but minor shade variations (slightly lighter/darker versions of reference colors, saturation differences within 15\%, 1-2 secondary elements use similar but non-exact colors)
            \item \textbf{Score 0}: Different color palette applied, incorrect categorical color assignments (e.g., Category A colored red instead of blue), reversed color scale (e.g., heatmap scale inverted), or 3+ elements with significantly different colors
        \end{itemize}
    \end{itemize}
    
    \item \textbf{Style and Format Consistency}
    \begin{itemize}
        \item \textbf{Question}: Does Chart B match Chart A's stylistic attributes (background color, grid styles, marker shapes, line patterns, font families, border styles)?
        \item \textbf{Scoring Criteria}:
        \begin{itemize}
            \item \textbf{Score 2}: All stylistic attributes match (grid style [solid/dashed/dotted], marker shapes, line widths, line patterns, font families, background colors, border styles, opacity levels)
            \item \textbf{Score 1}: Core visual style maintained with minor deviations (grid style slightly different [dashed vs dotted], marker shapes similar but not exact [circle vs hexagon], line width within ±1px, font family same category [all sans-serif] but different face)
            \item \textbf{Score 0}: Multiple style mismatches (wrong grid style + wrong markers + wrong font category), fundamentally different aesthetic (dark theme vs light theme), or critical style elements missing (no gridlines when reference has them, no borders when required)
        \end{itemize}
    \end{itemize}
    
    \item \textbf{Data Pattern Consistency}
    \begin{itemize}
        \item \textbf{Question}: Does Chart B visually replicate Chart A's data patterns (data point positions, trend shapes, distribution profiles, statistical markers)?
        \item \textbf{Scoring Criteria}:
        \begin{itemize}
            \item \textbf{Score 2}: Data patterns match reference with >95\% positional accuracy, all trend shapes replicated, distribution profiles visually identical, all statistical markers (means, medians, confidence intervals) present and correct
            \item \textbf{Score 1}: Core data trends and patterns recognizable (correlation directions correct, major peaks/troughs present, distribution shapes similar), but minor deviations (5-15\% positional variance, slight smoothing differences, 1-2 missing minor statistical markers, interpolation artifacts in small regions)
            \item \textbf{Score 0}: Data patterns significantly diverged (>15\% positional errors, trend directions incorrect, distribution shapes fundamentally different, major statistical markers missing/wrong), or substantial data missing/added
        \end{itemize}
    \end{itemize}
    
    \item \textbf{Component Completeness}
    \begin{itemize}
        \item \textbf{Scoring Criteria}:
        \begin{itemize}
            \item \textbf{Score 2}: Chart B contains all visual components from Chart A including all data series, markers, annotations, grid elements, legends, statistical overlays (trendlines, confidence intervals, reference lines), decorative features, and auxiliary elements
            \item \textbf{Score 1}: All primary components present (all main data series, essential legends, critical annotations, primary axis elements, key statistical markers) but missing 1-3 minor elements (optional secondary annotations, decorative styling features, non-essential reference lines, auxiliary data labels on individual points)
            \item \textbf{Score 0}: Missing critical components (1+ major data series absent, essential legend missing, key annotations omitted, primary axis elements incomplete, important statistical overlays missing) that significantly affect information completeness or interpretation accuracy
        \end{itemize}
    \end{itemize}
\end{enumerate}

\vspace{10pt}

\noindent\textbf{2. Execution Quality (3-Level Scoring: 0/1/2)} \\
Evaluate the technical execution quality of Chart B independently.

\vspace{5pt}

\begin{enumerate}
    \item \textbf{Visual Clarity}
    \begin{itemize}
        \item \textbf{Scoring Criteria}:
        \begin{itemize}
            \item \textbf{Score 2}: Zero overlap or occlusion affecting readability; all elements (subplots, titles, axis labels, legends, annotations, tick marks, data markers) are fully legible with adequate separation (minimum 3-5px spacing between interactive elements)
            \item \textbf{Score 1}: Minor overlap limited to 1-2 non-critical elements (legend box edge touching subplot border, secondary annotation slightly overlapping gridline, axis tick label near but not touching neighbor) that marginally affects aesthetics but not comprehension
            \item \textbf{Score 0}: Critical overlap affecting primary elements (title obscuring data, axis labels overlapping each other, legend blocking data points, tick marks colliding, data series indistinguishable due to occlusion) that severely impairs interpretation
        \end{itemize}
    \end{itemize}
    
    \item \textbf{Compositional Balance}
    \begin{itemize}
        \item \textbf{Scoring Criteria}:
        \begin{itemize}
            \item \textbf{Score 2}: Optimal spatial distribution with proportional element sizing (subplots balanced, margins uniform), harmonious spacing (consistent gaps between subplots, appropriate padding), effective whitespace utilization, professional aesthetic quality with clear visual hierarchy
            \item \textbf{Score 1}: Adequate composition with acceptable proportions and spacing, minor imbalances (one subplot slightly larger than others, margins vary by 5-10\%, spacing somewhat inconsistent) that don't significantly detract from usability or professionalism
            \item \textbf{Score 0}: Poor composition with disproportionate elements (subplots drastically different sizes without justification, margins vary by >20\%, cramped or excessive spacing, wasted whitespace, visual imbalance creating confusion about hierarchy or importance, unprofessional appearance
        \end{itemize}
    \end{itemize}
    
    \item \textbf{Typographic Quality}
    \begin{itemize}
        \item \textbf{Scoring Criteria}:
        \begin{itemize}
            \item \textbf{Score 2}: All text appropriately sized relative to chart dimensions (titles 14-18pt, axis labels 10-12pt, tick labels 8-10pt for standard charts), fully legible at standard viewing distance (arm's length for digital, 2-3ft for print), correctly spelled, semantically accurate, properly positioned without overlap
            \item \textbf{Score 1}: Text mostly appropriate with minor issues (1-2 labels slightly undersized [7-8pt where 10pt ideal] but still readable, minor positioning suboptimality [axis label slightly too far from axis], abbreviations used where full words preferred but meaning clear)
            \item \textbf{Score 0}: Text severely undersized (<7pt for body text) or oversized (>20pt for labels), illegible elements, spelling errors, semantic inaccuracies (wrong variable names, incorrect units), severely mispositioned labels (rotated when horizontal preferred, outside visible area)
        \end{itemize}
    \end{itemize}
\end{enumerate}

\vspace{10pt}

\noindent\textbf{Output Format:} \\
Your response \textbf{MUST} be a single, valid JSON object with no additional text. Use this exact structure:

\begin{verbatim}
{
  "visual_structure_alignment": {
    "chart_type_consistency": {
      "score": 1,
      "reason": "Detail Explanation"
    },
    "spatial_layout_consistency": {
      "score": 2,
      "reason": "Detail Explanation"
    },
    "text_element_consistency": {
      "score": 1,
      "reason": "Detail Explanation"
    },
    "axis_configuration_consistency": 
    { "score": 1,
      "reason": "Detail Explanation"
    },
    "color_scheme_consistency": {
      "score": 2,
      "reason": "Detail Explanation"
    },
    "style_and_format_consistency": {
      "score": 1,
      "reason": "Detail Explanation"
    },
    "data_pattern_consistency": {
      "score": 2,
      "reason": "Detail Explanation"
    },
    "component_completeness": {
      "score": 1,
      "reason": "Detail Explanation"
    }
  },
  "execution_quality": {
    "visual_clarity": {
      "score": 1,
      "reason": "Detail Explanation"
    },
    "compositional_balance": {
      "score": 2,
      "reason": "Detail Explanation"
    },
    "typographic_quality": {
      "score": 2,
      "reason": "Detail Explanation"
    }
  },
  "improvement_recommendations":
  "Priority fixes: (1) Add missing
  auxiliary bar overlay..."
}
\end{verbatim}

\end{tcolorbox}
\section{Additional Discussion}
\label{app:addtional_discussion}

\subsection{Performance Analysis Across Benchmarks}

\begin{figure}[t]
    \centering
    \includegraphics[width=\linewidth]{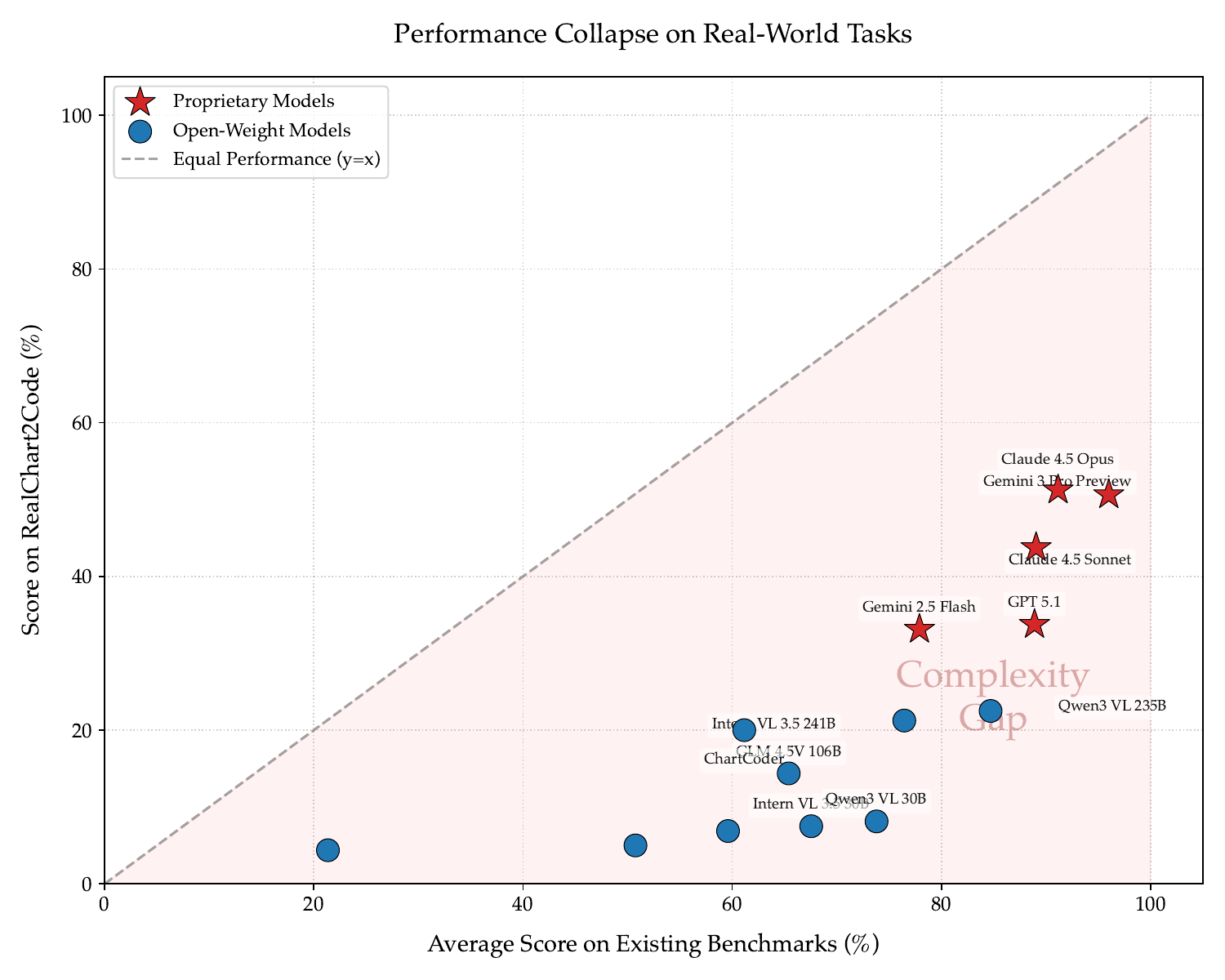}
    \caption{\textbf{Performance Collapse on Real-World Tasks.} We compare the normalized average scores (0-100\%) of models on existing benchmarks (x-axis) versus \texttt{RealChart2Code} (y-axis). The dashed line represents equal performance ($y=x$). The significant distance of all models from this line highlights the \textbf{Complexity Gap}: even state-of-the-art models that master simple plotting tasks see their performance drop by nearly 50\% when facing the realistic challenges in \texttt{RealChart2Code}.}
    \label{fig:benchmark_comparison}
\end{figure}

To contextualize the difficulty of \texttt{RealChart2Code}, we compare the performance of the evaluated models against two existing state-of-the-art benchmarks: \texttt{ChartMimic}~\citep{chartmimic} and \texttt{Plot2Code}~\citep{plot2code}. To ensure a fair comparison, we normalize all scores to a 0-100\% scale. As shown in Figure~\ref{fig:benchmark_comparison}, we observe a dramatic performance collapse when models transition from existing datasets to our proposed benchmark.

\paragraph{The "Complexity Gap".}
Existing benchmarks appear to have reached a saturation point for top-tier proprietary models. For instance, Gemini-3-Pro-Preview and Claude-4.5-Opus achieve near-perfect scores on existing tasks, averaging \textbf{96.0\%} and \textbf{91.2\%} respectively. However, their performance practically \textit{halves} on \texttt{RealChart2Code}, dropping to \textbf{50.6\%} and \textbf{51.3\%}. This vast disparity—visually represented by the "Complexity Gap" below the diagonal in Figure~\ref{fig:benchmark_comparison}—demonstrates that current SOTA models, while proficient at simple syntax, are far from solving the challenge of generating complex, multi-panel visualizations from authentic raw data.

\paragraph{Open-Weight Models Struggle.}
The gap is even more pronounced for open-weight models. While leading models like \texttt{Qwen3-VL-235B} perform admirably on standard benchmarks (averaging \textbf{84.7\%}), they fail to generalize to the rigor of real-world data science, scoring only \textbf{22.5\%} on \texttt{RealChart2Code}. This indicates that while open-weight models have memorized chart patterns, they lack the robust reasoning capabilities required for the iterative refinement and complex logic handling demanded by our benchmark.

\section{Case Study}
\label{app:case_study}

\begin{figure*}
    \centering
    \includegraphics[width=\linewidth]{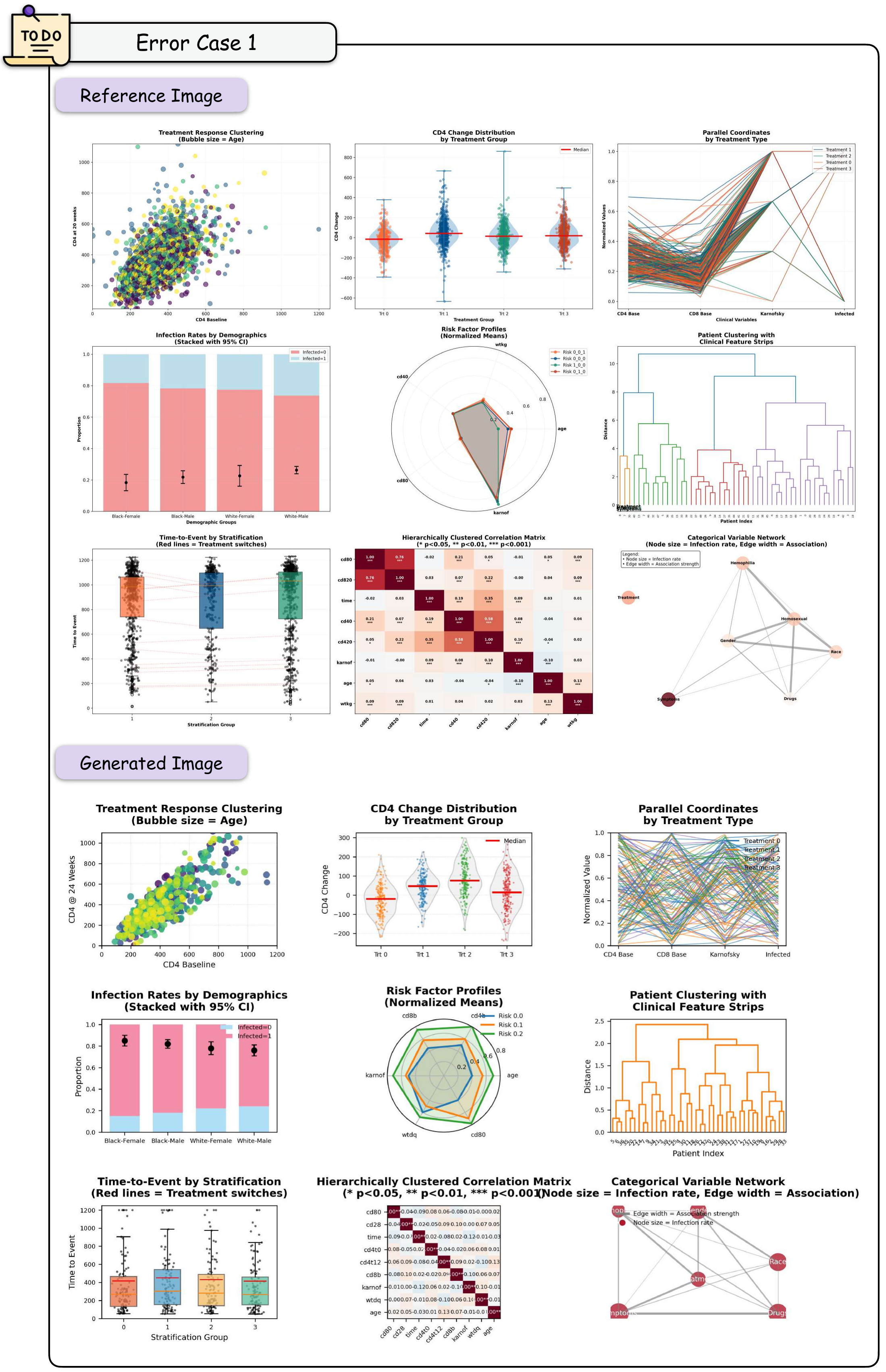}
\caption{\textbf{Error Case 1: Layout Optimization Failure.} Comparison between the reference visualization (top) and the model-generated output (bottom). While the generated code correctly implements the individual plotting logic for diverse chart types, it fails to utilize advanced layout managers (e.g., \texttt{GridSpec}), resulting in overcrowded subplots and overlapping elements.}
    \label{fig:error_case_1}
\end{figure*}

\paragraph{Error Case 1}
As shown in Figure \ref{fig:error_case_1}, the model demonstrates a strong capability in code generation for individual subplots, accurately replicating the visual style and data distribution of the reference. However, the synthesis of the multi-panel figure is structurally deficient. The generated script relies on default subplot positioning rather than explicit layout management tools such as \texttt{matplotlib.gridspec}. Consequently, the final output suffers from severe overcrowding, particularly in the bottom row where the text labels and plot elements overlap. This indicates that while the model captures local plotting semantics, it lacks the global spatial planning reasoning required for dense composite visualizations.

\begin{figure*}
    \centering
    \includegraphics[width=\linewidth]{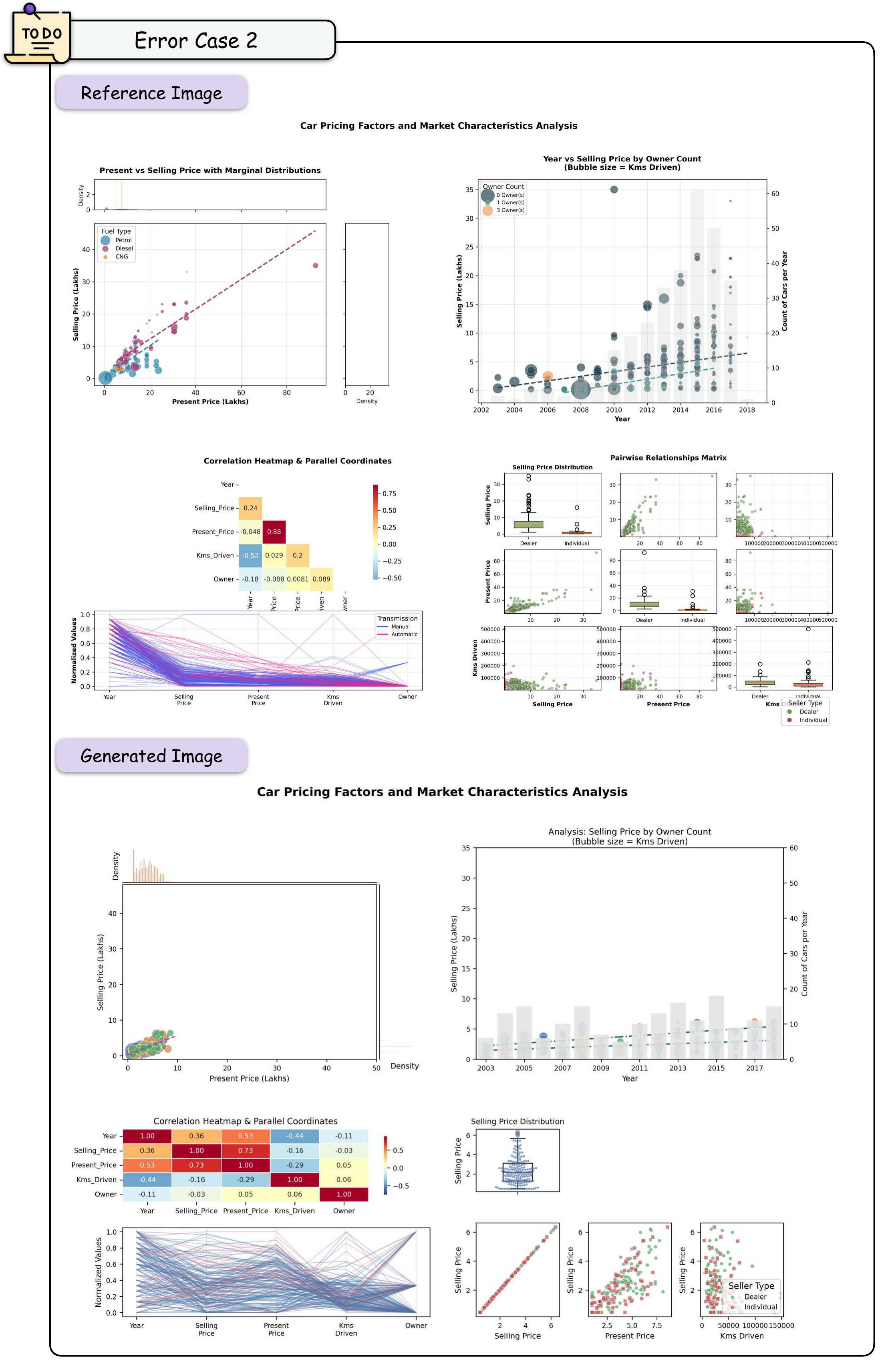}
    \caption{\textbf{Error Case 2: Complex Composite Layout Failure.} Comparison between the reference dashboard (top) and the model prediction (bottom). The target visualization features a highly intricate arrangement of heterogeneous charts, including marginal distributions, dual-axis overlays, and pairwise matrices. The model fails to recover the hierarchical structure, resulting in fragmented subplots, missing central data elements, and a collapse of the global layout.}
    \label{fig:error_case_2}

\end{figure*}

\paragraph{Error Case 2}
As illustrated in Figure \ref{fig:error_case_2}, this case pushes the boundaries of the model's spatial reasoning capabilities. The reference image requires nesting complex sub-layouts (e.g., a scatter matrix and a joint plot) within a larger dashboard grid. The model struggles significantly with this hierarchical composition. Specifically, the pairwise relationship matrix (bottom right) is rendered as disjointed, individual plots rather than a coherent grid, and the marginal joint plot (top left) fails to render the central scatter distribution entirely. This suggests that while the model can identify distinct plot types, it lacks the ability to orchestrate nested layout containers and coordinate systems required for high-density dashboards.

\begin{figure*}
    \centering
    \includegraphics[width=\linewidth]{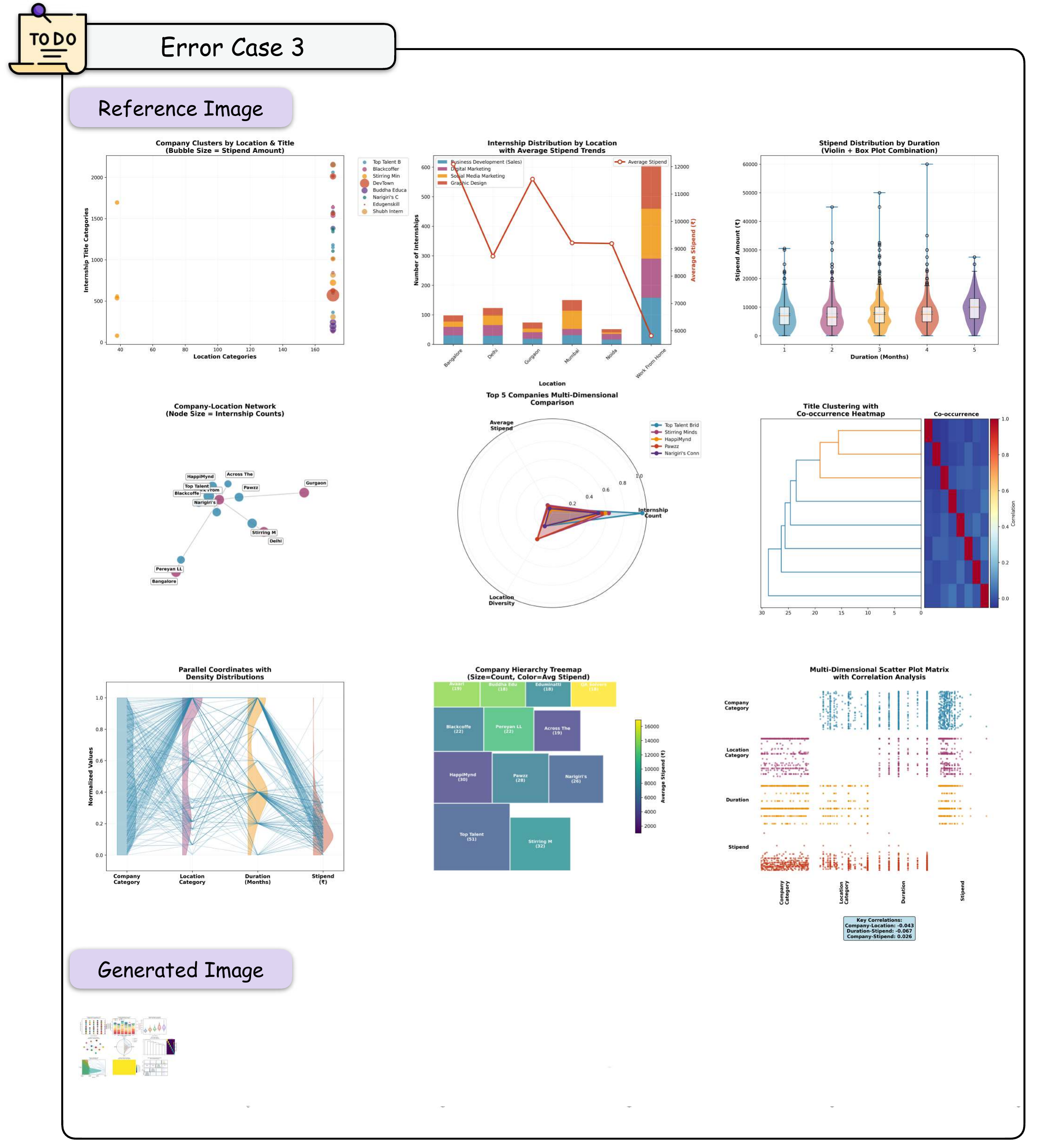}
    \caption{\textbf{Error Case 3: Canvas Scaling and Dimensioning Failure.} Comparison between the reference visualization (top) and the model-generated output (bottom). The model fails to correctly parametrize the global figure dimensions relative to the content, resulting in the entire subplot grid being compressed into a minute area. This scaling mismatch leaves the majority of the canvas as empty whitespace, rendering the visualization illegible.}
\label{fig:error_case_3}
\end{figure*}

\paragraph{Error Case 3}
As demonstrated in Figure \ref{fig:error_case_3}, the model exhibits a severe failure in global canvas utilization. While the code attempts to construct the multi-panel grid, it lacks the necessary logic to balance figure size (\texttt{figsize}) with element scaling and resolution. The generated plot suffers from a critical aspect ratio mismatch, where the visualization is rendered at a "thumbnail" scale within a disproportionately large canvas. This results in excessive whitespace on the right and top margins, indicating that the model struggles to align code parameters with standard visual output requirements for readability.

\paragraph{Error Case 4}

\begin{figure*}
    \centering
    \includegraphics[width=\linewidth]{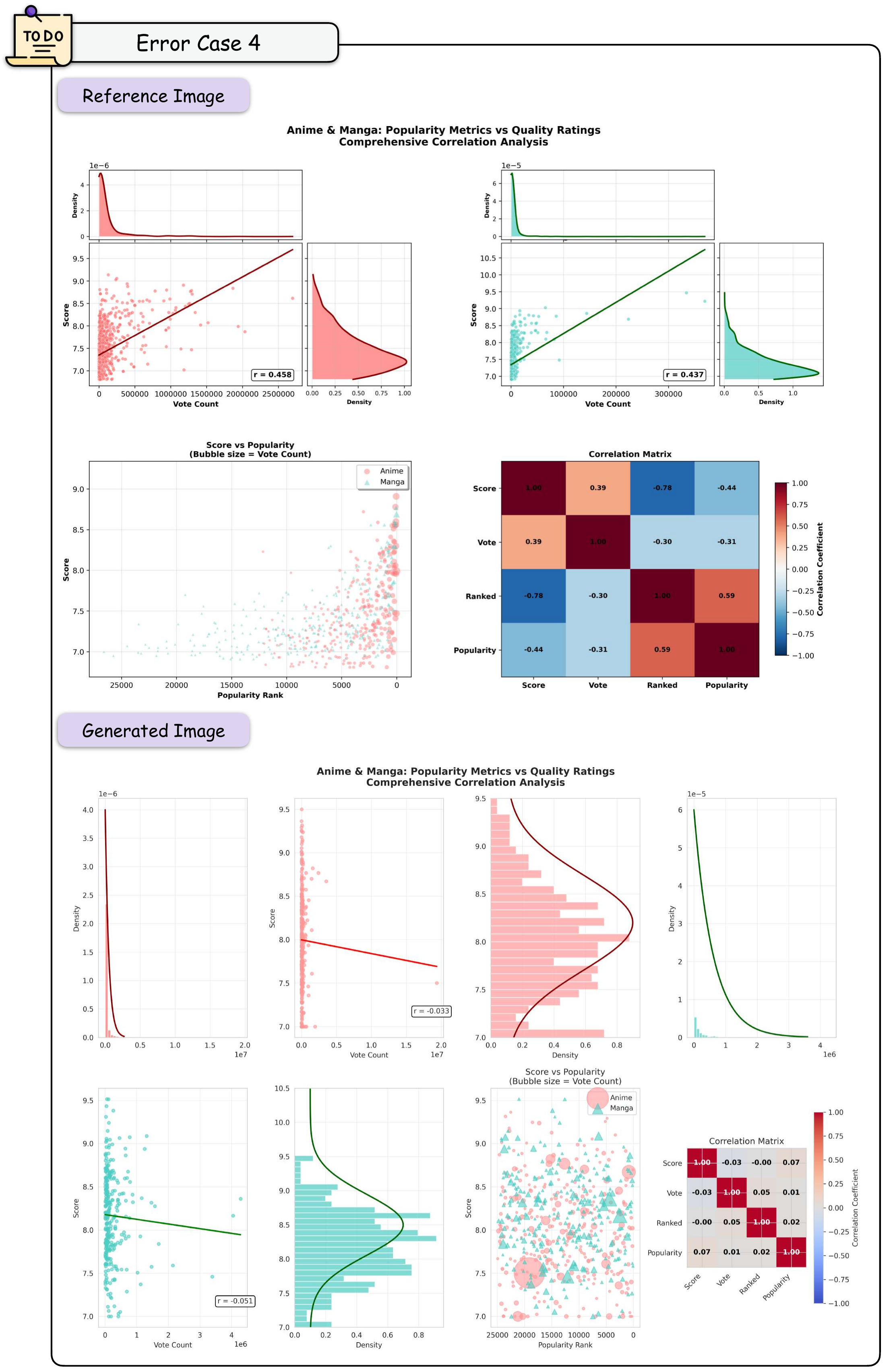}
    \caption{\textbf{Error Case 4: Semantic Layout Decomposition.} Comparison between the reference dashboard (top) and the model output (bottom). The reference relies on composite visualizations, specifically joint plots that integrate scatter plots with marginal density distributions. The model fails to perceive these grouped elements as unified semantic entities, instead breaking them down into disjointed, independent subplots. This results in a complete loss of the original grid structure and intended visual hierarchy.}
\label{fig:error_case_4}
\end{figure*}

Figure \ref{fig:error_case_4} illustrates a fundamental failure in parsing composite chart structures. The reference image features advanced "JointGrid" style layouts, where central axes are spatially locked with marginal axes to display bivariate distributions. The model, however, interprets these marginal components as standalone visualizations. Consequently, it "flattens" the hierarchical design, rendering the marginal histograms as separate, strictly grid-aligned panels rather than integrated overlays. This error demonstrates a lack of understanding regarding semantic grouping in data visualization, leading to a fragmented layout that fails to reconstruct the cohesive narrative of the original image.

\paragraph{Error Case 5}
Qwen3-VL-235B exhibits a recurring tendency to generate plausible-looking but functionally invalid API calls, affecting approximately 20\% of the generated solutions. Specifically, the model frequently executes \texttt{plt.style.use('seaborn-v0\_11')}, triggering a \texttt{ValueError} since this specific style parameter does not exist in standard Matplotlib or Seaborn libraries. This specific hallucination suggests a contamination or noise issue within the model's code-domain training data, where it has memorized incorrect versioning strings or deprecated syntax. Consequently, a significant portion of the outputs fail not due to visual reasoning deficits, but due to fundamental library knowledge errors.

\end{document}